\documentclass[journal]{IEEEtran}
\pdfoutput=1
\usepackage{graphicx}

\usepackage[OT1]{fontenc}

\usepackage{amsmath}
\usepackage{amsfonts}

\DeclareMathOperator*{\argmax}{arg\,max}

\usepackage{multirow}

\usepackage{algorithm}
\usepackage{algorithmic}
\usepackage{url}

\begin{document}
%
\title{Multimodal Emotion Recognition Using Deep Canonical
Correlation Analysis}
%
%
%

\author{Wei~Liu,
        Jie-Lin~Qiu, Wei-Long Zheng,~\IEEEmembership{Member, IEEE},  
and~Bao-Liang~Lu,~\IEEEmembership{Senior~Member,~IEEE}
\thanks{The work of Wei Liu, Jie-Lin Qiu, and Bao-Liang Lu was supported in part by the National Key Research and Development Program of China under Grant 2017YFB1002501, in part by the National Natural Science Foundation of China under Grant 61673266, in part by the Major Basic Research Program of Shanghai Science and Technology Committee under Grant 15JC1400103, in part by the ZBYY- MOE Joint Funding under Grant 6141A02022604, in part by the Technology Research and Development Program of China Railway Corporation under Grant 2016Z003-B, and in part by the Fundamental Research Funds for the Central Universities. (Corresponding author: Bao-Liang Lu.)}
\thanks{Wei Liu, and Bao-Liang Lu are with the Center for Brain-Like Computing and Machine Intelligence, Department of Computer Science and Engineering, Shanghai Jiao Tong University, Shanghai 200240, China, also with the Key Laboratory of Shanghai Education Commission for Intelligent Interaction and Cognitive Engineering, Shanghai Jiao Tong University, Shanghai 200240, China, and also with the Brain Science and Technology Research Center, Shanghai Jiao Tong University, Shanghai 200240, China (e-mail: blu@cs.sjtu.edu.cn).}
\thanks{Jie-Lin Qiu is with the Department of Electronic Engineering, Shanghai Jiao Tong University, Shanghai 200240, China.}
\thanks{Wei-Long Zheng is with the Department of Neurology, Massachusetts General Hospital, Harvard Medical School, Boston, MA 02114, USA.}
}
\maketitle

\begin{abstract}

Multimodal signals are more powerful than unimodal data for emotion recognition since they can represent emotions more comprehensively. 
In this paper, we introduce deep canonical correlation analysis 
(DCCA) to multimodal emotion recognition. 
The basic idea behind DCCA is to transform each modality separately and 
coordinate different modalities into a hyperspace by using specified canonical correlation analysis constraints. 
We evaluate the performance of DCCA on five multimodal datasets: the SEED, SEED-IV, SEED-V, DEAP, and DREAMER datasets. 
Our experimental results demonstrate that 
DCCA achieves state-of-the-art recognition accuracy rates on all five 
datasets: 94.58\% on the SEED dataset, 87.45\% on the SEED-IV dataset, 
84.33\% and 85.62\% for two binary classification tasks and 88.51\% for 
a four-category classification task on the DEAP dataset, 83.08\% on the
SEED-V dataset, and 88.99\%, 90.57\%, and 90.67\% for three binary 
classification tasks on the DREAMER dataset. 
We also compare the noise robustness of DCCA with that of existing methods when adding various amounts of noise to the SEED-V dataset. The experimental results indicate that DCCA has greater robustness.
By visualizing feature distributions with t-SNE and calculating the mutual information between different modalities before and after using DCCA, we find that the features transformed by DCCA from different modalities are more homogeneous and discriminative across emotions. 
\end{abstract}


%
\IEEEpeerreviewmaketitle

\section{Introduction}
%
%
%
%
\IEEEPARstart{E}{motion} strongly influences in our daily activities such as interactions between people, decision making, learning, and working. To endow a computer with emotion perception, understanding, and regulation abilities,
Picard {\it et al.} developed the concept of affective computing,
which aims to be used to study and develop systems and devices that can recognize, 
interpret, process, and simulate human 
affects~\cite{picard2000affective,picard2001toward}.
Human emotion recognition is a current hotspot in  affective computing research. 
Since emotion recognition is critical for applications such as affective brain-computer interaction, emotion regulation and the diagnosis of emotion-related diseases, it is necessary to build a reliable and accurate model for recognizing human emotions. 
\par 
Traditional emotion recognition systems are built with speech signals~\cite{el2011survey}, facial expressions~\cite{ko2018brief}, and non-physiological signals~\cite{yadollahi2017current}.
However, in addition to clues from external appearances, emotions contain reactions from the central and peripheral nervous systems.
Moreover, an obvious drawback of using behavioral modalities for emotion recognition is the uncertainty that arises in the case of individuals who either consciously regulate their emotional manifestations or are naturally suppressive.
In contrast, EEG-based emotion recognition has been proven to be a reliable method because of its high recognition accuracy, objective evaluation and stable neural patterns~\cite{zheng2015investigating,zheng2017identifying,yang2018eeg,yin2017cross}.
\par 
For the above reasons, researchers have 
tended to study emotions through physiological signals in recent years. These signals are more accurate and difficult to deliberately change by users.
Lin and colleagues evaluated music-induced emotion recognition with EEG signals and attempted to use as few electrodes as possible~\cite{lin2011generalizations}.
Wang and colleagues used EEG signals to classify positive and negative emotions and compared
different EEG features and classifiers~\cite{wang2014emotional}. 
Kim and Andr\'e showed that electromyogram, electrocardiogram, skin conductivity, and respiration changes were reliable signals for emotion recognition~\cite{kim2008}.
V{\~o} {\it et al.} studied the relationship between emotions and eye movement features, and they found that pupil diameters were influenced by 
both emotion and age~\cite{vo2008coupling}.
\par 
Emotions are complex cognitive processes that involve subjective experience, expressive behaviors, and psychophysiological changes.
Due to the rich characteristics of human emotions, it is difficult for single-modality signals to describe emotions comprehensively. Therefore, 
recognizing emotions with multiple modalities has become a promising method for building emotion recognition systems with high accuracy~\cite{poria2017,zheng2018emotionmeter,soleymani2012multimodal,6095505,8013713,7112127}.
Multimodal data can reflect emotional changes from multiple perspective, which is 
conducive to building a reliable and accurate emotion recognition model.\par 
Multimodal fusion is one of the key aspects in taking full advantage of multimodal signals. In the past few years, researchers have utilized various methods to fuse different modalities. Lu and colleagues employed feature-level concatenation, MAX fusion, SUM fusion, and fuzzy integral fusion to merge EEG and eye movement features, and they found the complementary properties of EEG and eye movement features in emotion recognition tasks~\cite{lu2015combining}. Koelstra and colleagues evaluated the feature-level concatenation of EEG features and peripheral physiological features, and they found that participant ratings and EEG frequencies were significantly correlated and that decision fusion achieved the best emotion recognition results~\cite{koelstra2012deap}. Sun {\it et al.} built a hierarchical classifier by combining both feature-level and decision-level fusion for emotion recognition tasks in the wild. The method was evaluated on several datasets and made very promising achievements on the validation and test sets~\cite{sun2016combining}.\par 
Currently, with the rapid development of deep learning, researchers are applying deep learning models to fuse multiple modalities.
Deep-learning-based multimodal representation frameworks
can be classified into two categories: multimodal joint representation and
multimodal coordinated representation~\cite{Baltru2017Multimodal}. Briefly,
the multimodal joint representation framework takes all the modalities as input, and each
modality starts with several individual neural layers followed by a hidden
layer that projects the modalities into a joint space.
The multimodal coordinated representation framework, instead of projecting the modalities
together into a joint space, learns separate representations for each modality
and coordinates them into a hyperspace with constraints between different modalities. 
Various multimodal joint representation frameworks have been applied to emotion 
recognition 
in very recent years~\cite{liu2016emotion,tang2017multimodal,li2016emotion,yin2017recognition}. 
However, the multimodal coordinated representation framework has not yet been fully studied.
\par 
In this paper, we introduce a coordinated representation model named Deep
Canonical Correlation Analysis (DCCA)~\cite{andrew2013deep,qiu2018multi} to multimodal emotion recognition. The basic idea behind DCCA is to learn separate but coordinated representations for each modality under canonical correlation analysis (CCA) constraints. Since the coordinated representations are of the same dimension, we denote the coordinated hyperspace by $\mathcal{S}$.\par 
Compared with our previous work~\cite{qiu2018multi},
the main contributions of this paper on multimodal emotion recognition can be summarized as follows:
\begin{itemize}
\item[1.] We introduce DCCA to multimodal emotion recognition and evaluate the effectiveness of DCCA on five benchmark datasets: the SEED, SEED-IV, SEED-V, DEAP, and DREAMER datasets. Our experimental results on these five datasets reveal that different emotions are disentangled in the coordinated hyperspace $\mathcal{S}$, and the transformation process of DCCA preserves emotion-related information and discards unrelated information.
\item[2.] We examine the robustness of DCCA and the existing methods on the SEED-V dataset under different levels of noise. The experimental results show that DCCA has higher robustness than the existing methods under most noise conditions.
\item[3.] By adjusting the weights of different modalities, DCCA allows users to fuse different modalities with greater flexibility such that various modalities contribute differently to the fused features.
\end{itemize}
\par 
The remainder of this paper is organized as follows. 
Section \ref{sec:related_work} summarizes the development and current state of 
multimodal fusion strategies. In Section 
\ref{sec:methods}, we introduce the algorithms for the canonical correlation analysis, DCCA, the baseline models utilized in this paper, and the mutual information neural estimation (MINE) algorithm. The experimental settings are reported in Section \ref{sec:exp}. Section \ref{sec:res} presents and analyzes the experimental results. Finally, conclusions are given in Section \ref{sec:conclusion}.

\section{Related Work}
\label{sec:related_work}
One of the key problems in multimodal deep learning is how to fuse data from different modalities. 
Multimodal fusion has gained increasing attention from researchers in diverse
fields due to its potential for innumerable applications such as emotion recognition, event detection, image segmentation, and video classification~\cite{lahat2015multimodal,D2015A}. 
According to the level of fusion, traditional fusion strategies can be classified into the following three categories: 1) feature-level fusion (early fusion), 2) decision-level multimodal fusion (late fusion), and 3) hybrid multimodal fusion. With the rapid development of deep learning, an increasing number of researchers are employing deep learning models to facilitate multimodal fusion. In the following, we introduce these multimodal fusion types and their subtypes.
\subsection{Feature-level fusion}
Feature-level fusion is a common and straightforward method to fuse different modalities. The features extracted from the various modalities are first combined into a high-dimensional feature and then sent as a whole to the models~\cite{hazarika2018self,koelstra2012deap,lu2015combining,ngiam2011multimodal,Monkaresi2012Classification}. \par 
The advantages of feature-level fusion are two-fold: 1) it can utilize the correlation between different modalities at an early stage, which better facilitates task accomplishment, and 2) the fused data contain more information than a single modality, and thus, a performance improvement is expected. The drawbacks of feature-level fusion methods mainly reside in the following: 1) it is difficult to represent the time synchronization between different modality features, 2) this type of fusion method might suffer the curse of dimensionality on small datasets, and 3) larger dimensional features might stress computational resources during model training.
\subsection{Decision-level fusion}
Decision-level fusion focuses on the usage of small classifiers and their combination. Ensemble learning is often used to assemble these classifiers. The term decision-level fusion describes a variety of methods designed to merge the outcomes and ensemble them into a single decision. \par 
Rule-based fusion methods are most adopted in multimodal emotion recognition. Lu and colleagues utilized MAX fusion, SUM fusion, and fuzzy integral fusion for multimodal emotion recognition, and they found the complementary nature of EEG and eye movement features by analyzing confusion matrices~\cite{lu2015combining}. Although rule-based fusion methods are easy to use, the difficulty facing rule-based fusion is how to design good rules. If rules are too simple, they might not reveal the relationships between different modalities.
\par 
 The advantage of decision-level fusion is that the decisions from different classifiers are easily compared and each modality can use its best suitable classifier for the task.
\subsection{Hybrid fusion}
Hybrid fusion is a combination of feature-level fusion and decision-level fusion. Sun and colleagues built a hierarchical classifier by combining both feature-level and decision-level fusion methods for emotion recognition~\cite{sun2016combining}. Guo {\it et al.} built a hybrid classifier by combining fuzzy cognitive map and SVM to classify emotional states with compressed sensing representation~\cite{guo2019}.

\subsection{Deep-learning-based fusion}
For deep learning models, different types of multimodal fusion methods have been developed, and
these methods can be grouped into two categories based on the
modality representation: multimodal joint representation and multimodal coordinated representation~\cite{Baltru2017Multimodal}.\par
The multimodal joint representation framework takes all the modalities as input, and each
modality starts with several individual neural layers followed by a hidden
layer that projects the modalities into a joint space. Both transformation and
fusion processes are achieved automatically by black-box models
and users do not know the meaning of the joint representations. The multimodal joint representation framework has been applied to
emotion recognition~\cite{liu2016emotion,tang2017multimodal} and natural language
processing~\cite{naim2014unsupervised}.\par
The multimodal coordinated representation framework, instead of projecting the modalities together into a joint space, learns separate representations for each modality but coordinates them through a constraint. 
The most common
coordinated representation models enforce similarity between modalities. Frome
and colleagues proposed a deep visual semantic embedding (DeViSE) model to
identify visual objects~\cite{frome2013devise}. DeViSE is initialized from two pre-trained neural network models: a visual object categorization network and a skip-gram language model. DeViSE combines these two networks by the dot-product and hinge rank loss similarity metrics such that the model is trained to produce a higher dot product similarity between the visual model output and the vector representation of the correct label than that between the visual output and other randomly chosen text terms.\par  
The deep canonical correlation analysis (DCCA) method, which is another model under the coordinated representation framework, was proposed by
Andrew and colleagues~\cite{andrew2013deep}. In contrast to DeViSE, DCCA adopts traditional CCA as a similarity metric, which allows us to transform data into a highly correlated hyper-space.
\section{Methods}
In this section, we first provide a brief description of traditional canonical correlation analysis (CCA) in Section \ref{subsec:cca}. Based on CCA, we present the building processes of DCCA in Section \ref{subsec:dcca}. The baseline methods used in this paper are described in Section~\ref{subsec:baseline}. Finally, the mutual information neural estimation (MINE) algorithm is given in Section \ref{subsec:mine}, which is utilized to analyze the properties of transformed features implemented by DCCA in the coordinated hyperspace $\mathcal{S}$.
\label{sec:methods}
\subsection{Canonical Correlation Analysis}
\label{subsec:cca}
Canonical correlation analysis (CCA) was proposed by
Hotelling~\cite{hotelling1992relations}. It is a widely used technique in the
statistics community to measure the linear relationship between two
multidimensional variables. Hardoon and colleagues applied CCA to machine
learning~\cite{hardoon2004canonical}.\par
Let $(X_1, X_2)\in \mathbb{R}^{n_1}\times \mathbb{R}^{n_2}$ denote random
vectors with covariance matrices $(\Sigma_{11}, \Sigma_{22})$ and
cross-variance matrix $\Sigma_{12}$. CCA attempts to find linear transformations
of $(X_1, X_2)$, $(w_1^*X_1, w_2^*X_2)$, which are maximally correlated:
\begin{align}
	(w_1^*, w_2^*) &= \nonumber\argmax_{w_1, w_2}\ corr(w_1'X_1, w_2'X_2)\\ 
& =\argmax_{w_1,
w_2}\frac{w_1'\Sigma_{12}w_2}{\sqrt{w_1'\Sigma_{11}w_1w_2'\Sigma_{22}w_2}
}\label{eq:cca_raw}.
\end{align}
Since Eq. (\ref{eq:cca_raw}) is invariant to the scaling of the weights $w_1$ and $w_2$, Eq. (\ref{eq:cca_raw}) can be reformulated as follows:
\begin{equation}
(w_1^*, w_2^*) =
\argmax_{w_1'\Sigma_{11}w_1=w_2'\Sigma_{22}w_2=1}w_1'\Sigma_{12}w_2,
\end{equation}
where we assume the projections are constrained to have unit variance.\par 
To find multiple results of $(w_1^i, w_2^i)$, subsequent projections are
also constrained to be uncorrelated with previous projections, {\it i.e.},
$w_1^i\Sigma_{11}w_1^j=w_2^i\Sigma_{22}w_2^j=0$ for $i<j$. Combining the top
$k$ projection vectors $w_1^i$ into a matrix $A_1\in \mathbb{R}^{n_1\times k}$
as column vectors and similarly placing $w_2^i$ into
$A_2\in\mathbb{R}^{n_2\times k}$, we then identify the top $k \leq
\min(n_1,n_2)$ projections:
\begin{align}
	\nonumber\text{maximize:}&\ \ tr(A_1'\Sigma_{12}A_2)\\
	\text{subject to:}&\ \ A_1'\Sigma_{11}A_1 = A_2'\Sigma_{22}A_2 = I.
\end{align}
\par 
To solve this objective function, 
we first define
$T=\Sigma_{11}^{-1/2}\Sigma_{12}\Sigma_{22}^{-1/2}$, and we let $U_k$ and $V_k$ be
the matrices of the first $k$ left singular and right singular vectors of $T$, respectively.
Then the optimal objective value is the sum of the top $k$ singular values of
$T$, and the optimum is obtained at $(A_1^*, A_2^*)=(\Sigma_{11}^{-1/2}U_k,
\Sigma_{22}^{-1/2}V_k)$. This method requires the covariance matrices
$\Sigma_{11}$ and $\Sigma_{22}$ to be nonsingular, which is usually satisfied in
practice.\par
For the original CCA, the representations in the latent space are obtained by linear transformations, which limit the scope of application of CCA. To address this problem, Lai and Fyfe~\cite{lai2000kernel} proposed kernel CCA, in which kernel methods are introduced for nonlinear transformations. Klami and colleagues developed probabilistic canonical correlation analysis (PCCA)~\cite{Klami2008Probabilistic}; later, they extended PCCA to a Bayesian-based CCA named inter-battery factor analysis~\cite{klami2013bayesian}. There are many other extensions of CCA such as tensor CCA~\cite{kim2007tensor}, sparse CCA~\cite{hardoon2011sparse}, and cluster CCA~\cite{rasiwasia2014cluster}.

\subsection{Deep Canonical Correlation Analysis}
\label{subsec:dcca}
\begin{figure}[ht]
\centering 
\includegraphics[width=.35\textwidth]{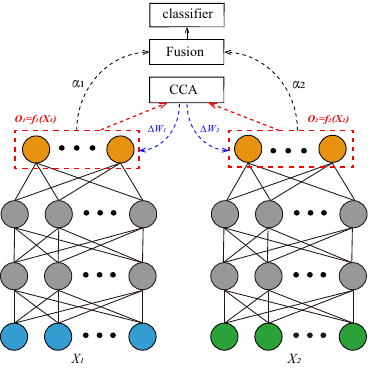}
\caption{The structure of DCCA. Different modalities are transformed by different neural networks separately. The outputs ($O_1,O_2$) are regularized by the traditional CCA constraint. Various strategies can be adopted to fuse $O_1$ and $O_2$, and we use the weighted sum fusion method as shown in Eq. (\ref{eq:weighted_sum_fusion}). We update the parameters to maximize the CCA metric of different modalities, and the fused features are used to train a classifier.}
\label{fig:dcca_structure}
\end{figure}

In this paper, we introduce deep canonical correlation analysis (DCCA) to multimodal emotion recognition. 
DCCA was proposed
by Andrew and colleagues~\cite{andrew2013deep}, and it computes representations of multiple
modalities
by passing them through multiple stacked layers of nonlinear transformations.
Figure~\ref{fig:dcca_structure} depicts the structure of DCCA used in this paper.\par 
Let $X_1\in\mathbb{R}^{N\times d_1}$ be the instance matrix for the first
modality and
$X_2\in\mathbb{R}^{N\times d_2}$ be the instance matrix for the second modality. Here, $N$ is the
number of instances, and $d_1$ and $d_2$ are the dimensions of the extracted features for these two modalities, respectively.
To transform the raw features of two modalities nonlinearly, we build two deep neural networks for the two modalities as follows:
\begin{align}
O_1 =& f_1(X_1;W_1), \\
O_2 =& f_2(X_2;W_2),
\end{align}
where $W_1$ and $W_2$ denote all parameters for the non-linear transformations, 
$O_1\in\mathbb{R}^{N\times d}$ and $O_2\in\mathbb{R}^{N\times d}$ are the
outputs of the neural networks, and $d$ denotes the output dimension of DCCA.
The goal of DCCA is to jointly learn the parameters $W_1$ and $W_2$ for both 
neural networks such that the correlation of $O_1$ and $O_2$ is as high as
possible:
\begin{equation}
(W_1^*, W_2^*) = \argmax_{W_1, W_2}\ corr(f_1(X_1;W_1), F_2(X_2;W_2)).
\label{eq:objective}
\end{equation}
\par 
We use the backpropagation algorithm to update $W_1$ and $W_2$. The solution to calculating the gradients of the objective function in Eq. 
(\ref{eq:objective}) was developed by Andrew and colleagues~\cite{andrew2013deep}.
Let $\bar{O}_1 = O_1'-\frac{1}{N}O_1'\mathbf{1}$ be the centered output matrix
(similar to $\bar{O}_2$).
We define $\hat{\Sigma}_{12}=\frac{1}{N-1}\bar{O}_1\bar{O}_2'$, 
$\hat{\Sigma}_{11}=\frac{1}{N-1}\bar{O}_1\bar{O}_1'+r_1\mathbf{I}$. Here, $r_1$ is a regularization constant (similar to $\hat{\Sigma}_{22}$). The total correlation of the top $k$ components of $O_1$ and $O_2$ is the sum of the top $k$ singular values of matrix $T =
\hat{\Sigma}_{11}^{-1/2}\hat{\Sigma}_{12}\hat{\Sigma}_{22}^{-1/2}$. In this paper, we take $k=d$, and the total correlation is the trace of $T$:
\begin{equation}
corr(O_1, O_2) = \bigg(tr(T'T)\bigg)^{1/2}.
\end{equation}
\par 
Finally, we calculate the gradients with the singular decomposition of
$T=UDV'$,
\begin{equation}
\frac{\partial corr(O_1, O_2)}{\partial O_1} =
\frac{1}{N-1}(2\nabla_{11}\bar{O}_1+\nabla_{12}\bar{O}_2),
\end{equation}
where 
\begin{align}
\nabla_{11} =&
-\frac{1}{2}\hat{\Sigma}_{11}^{-1/2}UDU'\hat{\Sigma}_{11}^{-1/2},\\
\nabla_{12} =& \hat{\Sigma}_{11}^{-1/2}UV'\hat{\Sigma}_{22}^{-1/2},
\end{align}
and $\partial corr(O_1,O_2)/\partial O_2$ has a symmetric expression.\par 
After the training of the two neural networks, the transformed features $O_1, O_2\in\mathcal{S}$ are in the coordinated hyperspace $\mathcal{S}$. 
In the original DCCA~\cite{andrew2013deep}, the authors did not explicitly describe how to use transformed features for real-world applications via machine learning algorithms. Users need to design a strategy to take advantage of the transformed features according to their application.\par 
 In this paper, we use a weighted sum fusion method to obtain the fused features as follows:
 \begin{equation}
 	O=\alpha_1O_1+\alpha_2O_2,
 	\label{eq:weighted_sum_fusion}
 \end{equation}
 where $\alpha_1$ and $\alpha_2$ are weights satisfying $\alpha_1 + \alpha_2=1$.
The fused features $O$ are used to train the classifiers
to recognize different emotions. In this paper, an SVM classifier is adopted.\par 
According to the construction processes mentioned above, DCCA brings the following advantages to multimodal emotion recognition:
\begin{itemize}
	\item By transforming different modalities separately, we can explicitly extract transformed features for each modality ($O_1$ and $O_2$) so that it is convenient to examine the characteristics and relationships of modality-specific transformations.
	\item With specified CCA constraints, we can regulate the non-linear mappings ($f_1(\cdot)$ and $f_2(\cdot)$) and make the model preserve  the emotion-related information.
	\item By using a weighted sum fusion (under the condition $\alpha_1+\alpha_2=1$), we can assign different priorities to these modalities based on our priori knowledge. A larger weight represents a larger contribution of the corresponding modality to the fusion features.
\end{itemize}
\subsection{Baseline methods}
\label{subsec:baseline}
\subsubsection{Concatenation Fusion}
The concatenation fusion is a type of feature-level fusion. The feature vectors
from two modalities are denoted as $X^1=[x^1_1,\cdots,x^1_n]\in\mathcal{R}^n$ and
$X^2=[x^2_1,\cdots,x^2_m]\in\mathcal{R}^m$, and the fused features can be calculated
with the following equation:
\begin{align}
	\nonumber X_{fusion} &= Concat([X^1, X^2]) \\
	&= [x^1_1,\cdots,x^1_n, x^2_1,\cdots,x^2_m].
\end{align}
\subsubsection{MAX Fusion}
The MAX fusion method is a type of decision-level fusion method that
chooses the class of the maximum probability as the prediction
result. Assuming that we have $K$ classifiers and $C$ categories, there is a 
probability distribution for each sample $P_j(Y_i|x_t), j\in\{1,\cdots,K\}$, and $i\in\{1,\cdots, C\}$, where $x_t$ is a sample, $Y_i$ is the predicted label, 
and $P_j(Y_i|x_t)$ is the probability of sample $x_t$ belonging to class $i$ generated by the $j$-th classifier.
The MAX fusion rule can be expressed as follows:
\begin{equation}
	\hat{Y} = \argmax_i\{\argmax_j P_j(Y_i|x_t)\}.
\end{equation}

\subsubsection{Fuzzy Integral Fusion}
The fuzzy integral fusion is also a type of decision-level fusion~\cite{Grabisch2000Application,li2012gender}. 
A fuzzy measure $\mu$ on the set $X$ is a function: $\mu: \mathcal{P}(X)\rightarrow [0,1]$, which satisfies the two axioms: 1) $\mu(\emptyset)=0$ and 2) $A\subset B\subset X \ \text{implies}\ \mu(A)\leq\mu(B)$. In this paper, we use the discrete
Choquet integral to fuse the multimodal features. The discrete Choquet integral of a
function $f:X\rightarrow \mathcal{R}^+$ with respect to $\mu$ is defined by
\begin{equation}
	\mathcal{C}_\mu(f):=\sum_{i=1}^{n}\big(f(x_{i})-f(x_{i-1})\big)\mu(A_{(i)}),
\end{equation}
where $\cdot_{(i)}$ indicates that the indices have been 
permuted such that $0\leq f(x_{(1)})\leq\cdots\leq f(x_{(n)})$, 
$A_{(i)}:=\{x_{(i)},\cdots,x_{(n)}\}$,
and $f(x_{(0)})=0$.\par 
In this paper, we utilize the algorithm proposed by Tanaka and Sugeno~\cite{Tanaka1991A} to calculate the fuzzy measure. The algorithm attempts to find the fuzzy measure $\mu$, which minimizes the total squared error of the model. Tanaka and Sugeno proved that the minimization problem can be solved through a quadratic programming method.
\subsubsection{Bimodal Deep AutoEncoder (BDAE)}
BDAE was proposed by Ngiam and colleagues~\cite{ngiam2011multimodal}. In our previous work, we applied BDAE to multimodal emotion recognition~\cite{liu2016emotion}.
\par 
A building block of BDAE is the restricted Boltzmann machine (RBM). 
The RBM is an undirected graph model, which has a visible
layer and a hidden layer. Connections exist only between the visible layer and hidden layer
, and there are no connections in the visible layer or in the hidden layer.
In this paper, we adopted the {\it BernoulliRBM} in Scikit-learn\footnote{\url{https://scikit-learn.org/stable/modules/generated/sklearn.neural_network.BernoulliRBM.html}}~\cite{scikit-learn}. 
The visible variables $\mathbf{v}\sim \text{Bern}(p)$ are binary stochastic units of dimension $M$, which means that the input data should be either binary or real valued between 0 and 1, signifying the probability.
The hidden variables also satisfy a Bernoulli distribution $\mathbf{h}\in\{0,1\}^N$.
The energy is calculated with the following function:
\begin{equation}
  \label{equ:energy}
  E(\mathbf{v},\mathbf{h};\theta) = -\sum_{i=1}^M\sum_{j=1}^NW_{ij}v_ih_j-\sum_{i=1}^M
  b_iv_i - \sum_{j=1}^Na_jh_j,
\end{equation}
where $\theta = \{\mathbf{a,b,W}\}$ are parameters, $W_{ij}$ is the symmetric weight
between the visible unit $i$ and the hidden unit $j$, and $b_i$ and $a_j$ are the bias terms of the visible
unit and hidden unit, respectively. With an energy function, we can obtain the joint
distribution over the visible and hidden units:
\begin{align}
  \label{equ:joint}
  &p(\mathbf{v},\mathbf{h};\theta) = \frac{1}{\mathcal{Z}(\theta)}\exp(E(\mathbf{v},\mathbf{h};\theta)), \\
  &\mathcal{Z}(\theta) = \sum_\mathbf{v}\sum_\mathbf{h}\exp(E(\mathbf{v},\mathbf{h};\theta)),
\end{align}
where $\mathcal{Z}(\theta)$ is the normalization constant.
Given a set of visible variables $\{\mathbf{v}_n\}_{n=1}^N$, the derivative of  the log-likelihood
with respect to the weight $\mathbf{W}$ can be calculated from Eq. (\ref{equ:joint}):
\begin{equation}
\frac{1}{N}\sum_{i=1}^N\frac{\partial \log p(\mathbf{v}_n;\theta)}{\partial W_{ij}}
 = \mathbb{E}_{P_{data}}[v_ih_j]-\mathbb{E}_{P_{model}}[v_ih_j].
\end{equation}
\par 
The BDAE training procedure includes encoding and decoding. 
In the encoding phase, we train two RBMs for EEG features and eye movement features, and the hidden layers are denoted as $h_{EEG}$ and $h_{Eye}$.
These two hidden layers are concatenated together, and the concatenated layer is used as the visual layer of a new upper RBM.
In the decoding stage, we unfold the stacked RBMs to reconstruct the input features. Finally, we use a back-propagation algorithm to minimize the reconstruction error.

\subsection{Mutual Information Neural Estimation}
\label{subsec:mine}
Mutual information is a fundamental quantity for measuring the relationship
between variables. 
The mutual information quantifies the dependence of two random variables
$X$ and $Z$ with the following equation:
\begin{equation}
	I(X;Z)=\int_{\mathcal{X}\times\mathcal{Z}}\log\frac{d\mathbb{P}_{XZ}}{d\mathbb{P}_{X}\otimes\mathbb{P}_{Z}}d\mathbb{P}_{XZ},
\end{equation}
where $\mathbb{P}_{XZ}$ is the joint probability distribution, and $\mathbb{P}_X=\int_{Z}d\mathbb{P}_{XZ}$ and $\mathbb{P}_Z=\int_{X}d\mathbb{P}_{XZ}$ are 
the marginals.\par 
The mutual information neural estimation (MINE) was proposed by Belghazi and 
colleagues~\cite{belghazi2018mine}. MINE is linearly scalable in dimensionality as well as in sample size, trainable through a back-propagation algorithm, and strongly consistent.\par 
The idea behind MINE is to choose $\mathcal{F}$ to be the family of functions
$T_\theta:\mathcal{X}\times\mathcal{Z}\to \mathbb{R}$ parameterized by a deep
neural network with parameters $\theta\in\Theta$. Then, the 
deep neural network is used to update the estimated mutual information,
\begin{equation}
	I(X;Z)\geq I_\Theta(X;Z),
\end{equation}
where $I_\Theta$ is defined as
\begin{equation}
\label{eq:mi}
	I_\Theta(X;Z)= \sup_{\theta\in\Theta}\mathbb{E}_{\mathbb{P}_{XZ}}[T_\theta]-\log(\mathbb{E}_{\mathbb{P}_X\otimes\mathbb{P}_Z}[e^{T_\theta}]).
\end{equation}
The expectations in Eq. (\ref{eq:mi}) are estimated using empirical samples
from $\mathbb{P}_{XZ}$ and $\mathbb{P}_X\otimes\mathbb{P}_Z$ or by shuffling the 
samples from the joint distribution, and MINE is defined as
\begin{equation}
	\widehat{I(X;Z)}_n=\sup_{\theta\in\Theta}\mathbb{E}_{\mathbb{P}^{(n)}_{XZ}}[T_\theta]-\log(\mathbb{E}_{\mathbb{P}^{(n)}_X\otimes\hat{\mathbb{P}}^{(n)}_Z}[e^{T_\theta}]),
\end{equation}
where $\hat{\mathbb{P}}$ is the empirical distribution associated with $n\ i.i.d.$ samples. The details on the implementation of MINE are provided in Algorithm \ref{alg:mine}.

\begin{algorithm}
\caption{Mutual Information Calculation between Two Modalities with MINE}
\label{alg:mine}
\begin{algorithmic}
\STATE {Input: Features extracted from two modalities:}
\STATE {\quad\quad\quad\  $X=\{x^1,\cdots, x^n\}$, $Z=\{z^1,\cdots, z^n\}$}
\STATE {Output: Estimated mutual information}
\STATE {$\theta\gets$initialize network parameters} 
\REPEAT 
\STATE {1. Draw $b$ mini-batch samples from the joint distribution:}
\STATE {\quad $(x^{(1)},z^{(1)}),\dots,(x^{(b)},z^{(b)})\sim\mathbb{P}_{XZ}$}
\STATE {2. Draw $b$ samples from the $Z$ marginal distribution:}
\STATE {\quad $\bar{z}^{(1)},\dots,\bar{z}^{(b)}\sim\mathbb{P}_Z$}
\STATE 3. Evaluate the lower bound:
\STATE {\quad $\mathcal{V}(\theta)\gets\frac{1}{b}\sum T_\theta(x^{(i)},z^{(i)})-log(\frac{1}{b}\sum e^{T_\theta(x^{(i)},z^{(i)})})$}
\STATE 4. Evaluate bias-corrected gradients:
\STATE {\quad $\hat{G}(\theta)\gets\tilde{\nabla}_\theta\mathcal{V}(\theta)$}
\STATE 5. Update the deep neural network parameters:
\STATE {\quad $\theta \gets \theta + \hat{G}(\theta)$}
\UNTIL{Convergence} 
\end{algorithmic}
\end{algorithm}
We modify the code of the MINE algorithm written by Masanori Yamada\footnote{\url{https://github.com/MasanoriYamada/Mine_pytorch}}; the code used in this paper can be downloaded from GitHub\footnote{\url{https://github.com/csliuwei/MI_plot}}.

\section{Experimental settings}
\label{sec:exp}
\subsection{Datasets}
To evaluate the effectiveness of DCCA for multimodal emotion recognition, five multimodal emotion recognition datasets are selected for experimental study in this paper.
\subsubsection{SEED dataset\protect\footnote{\protect\url{http://bcmi.sjtu.edu.cn/home/seed/index.html}}} 
The SEED dataset was developed by Zheng and Lu~\cite{zheng2015investigating}. A total of 15 Chinese film clips of three emotions (happy, neutral and sad) were chosen from a pool of materials as stimuli used in the experiments. Before the experiments, the participants were told the procedures of the entire experiment. During the experiments, the participants were asked to watch the selected 15 movie clips, and report their emotional feelings. After watching a movie clip, the subjects were given 45 seconds to provide feedback and 15 seconds to rest. In this paper, we use the same subset of the SEED dataset as in our previous work~\cite{lu2015combining,liu2016emotion,tang2017multimodal} for the comparison study.\par 
The SEED dataset contains EEG signals and eye movement signals. The EEG signals were collected with an ESI NeuroScan system at a sampling rate of 1000~Hz from a 62-channel electrode cap. Eye movement signals were collected with SMI eye tracking glasses\footnote{\url{https://www.smivision.com/eye-tracking/product/eye-tracking-glasses/}}.
\subsubsection{SEED-IV dataset}
The SEED-IV dataset was first proposed in ~\cite{zheng2018emotionmeter}. The experimental procedure was similar to that of the SEED dataset, and 72 film clips were chosen as stimuli materials. 
The dataset contains emotional EEG signals and eye movement signals of four 
different emotions, {\it i.e.}, happy, sad, neutral, and fear. 
A total of 15 subjects (7 male and 8 female) participated in the
experiments.
For each participant, three sessions were performed on different days, and each
session
consisted of 24 trials. In each trial, the participant watched one of the movie
clips.
\subsubsection{SEED-V dataset}
The SEED-V dataset was proposed in~\cite{li2019classification}. The dataset
contains EEG signals and eye movement signals for five emotions
(happy, sad, neutral, fear, and disgust). A total of 16 subjects (6 male and 10 female) were recruited to 
participate in the experiment, and each of them
performed the experiment three times. During the experiment, the subject were
required to watch 15 movie clips (3 clips for each emotion).
The same devices were used in the SEED-V dataset as in the SEED and
SEED-IV datasets. The SEED-V dataset used in this paper will be freely available to the academic community as a subset of SEED\footnote{\url{http://bcmi.sjtu.edu.cn/home/seed/index.html}}.
\subsubsection{DEAP dataset}
The DEAP dataset was developed by Koelstra and
colleagues~\cite{koelstra2012deap} and is a multimodal dataset for the
analysis of human affective states.
The EEG signals and peripheral physiological signals (EOG, EMG, GSR,
respiration belt, and plethysmograph) of 32 participants were recorded as each
watched
40 one-minute long excerpts of music videos. Participants rated each video in
terms of the levels of arousal, valence, like/dislike, dominance, and
familiarity.

\subsubsection{DREAMER dataset}
The DREAMER dataset is a multimodal emotion dataset developed by Katsigiannis and Ramzan~\cite{katsigiannis2017dreamer}. The DREAMER dataset consists of 14-channel EEG signals and 2-channel ECG signals of 23 subjects (14 males and 9 females).  During the experiments, the participants watched 18 film clips to elicit 9 different emotions including amusement, excitement, happiness, calmness, anger, disgust, fear, sadness, and surprise. After watching a clip, the self-assessment manikins were used to acquire subjective assessments of valence, arousal, and dominance. \par
\subsection{Feature extraction}
\subsubsection{EEG feature extraction}
For EEG signals, we extract differential entropy (DE)
features using short-term Fourier transforms with a 4-second Hanning window without overlapping~\cite{duan2013differential,shi2013differential}.
The differential entropy feature is used to measure the complexity of continuous random variables. 
Its calculation formula can be written as follows:
\begin{equation}
	h(\mathbf{X})=-\int_\mathbf{X}f(x)\log\big(f(x)\big)dx,
\end{equation}
where $\mathbf{X}$ is a random variable and $f(x)$ is the probability density
function of $\mathbf{X}$. For the time series $\mathbf{X}$, obeying the Gauss
distribution $N(\mu,\sigma^2)$, its differential entropy can be calculated as
follows:
\begin{align}
\nonumber h(\mathbf{X})&=-\int\limits_{-\infty}^{\infty}\frac{1}{\sqrt{2\pi\sigma^2}}e^{
-\frac{(x-\mu)^2}{2\sigma^2}}\log\Big({\frac{1}{\sqrt{2\pi\sigma^2}}}e^{-\frac{
(x-\mu)^2}{2\sigma^2}}\Big)dx\\
	&=\frac{1}{2}\log2\pi e\sigma^2.
\label{eq:de_cal}
\end{align}
\par 
Shi and colleagues~\cite{shi2013differential} proved that EEG signals within a short time period in different frequency bands are
subject to a Gaussian distribution by the Kolmogorov-Smirnov test, and the
DE features can be calculated by Eq.~(\ref{eq:de_cal}).\par 
We extract DE features from EEG signals (from the SEED, SEED-IV and SEED-V datasets) in five frequency bands for all channels: delta (1-4~Hz), theta (4-8~Hz), alpha (8-14~Hz), beta (14-31~Hz), and gamma (31-50~Hz). 
There are in total $62\times 5=310$ dimensions for 62 EEG
channels. 
Finally we adopt the linear dynamic system method to filter out noise and artifacts~\cite{shi2010off}.\par
For the DEAP dataset, the raw EEG signals were downsampled to 128 Hz and
preprocessed with a bandpass filter
from 4 to 75 Hz. We extract the DE features from four frequency bands
(theta, alpha, beta, and
gamma). As a result, there are 128 dimensions for the DE features.
\subsubsection{ECG feature extraction}
In previous ECG-based emotion recognition studies, researchers extracted time-domain features, frequency-domain features, and time-frequency-domain features from ECG signals for emotion recognition~\cite{katsigiannis2017dreamer,8219396,zhao2016wireless}. 
Katsigiannis and Ramzan extracted power spectral density (PSD) features of low frequency and high frequency from ECG signals~\cite{katsigiannis2017dreamer}. Hsu and colleagues extracted power for three frequency bands: a very-low-frequency range (0.0033 -- 0.04 Hz), a low-frequency range (0.04 -- 0.15 Hz), and a high-frequency range (0.15 -- 0.4 Hz)~\cite{8219396}.\par 
However, previous studies have shown that ECG signals have a much wider frequency range. In the early stage of ECG research, Scher and Young showed that ECG signals contain frequency components as high as 100 Hz~\cite{scher1960frequency}. Recently, Shufni and Mashor also showed that there are high-frequency components (up to 600 Hz) in ECG signals~\cite{shufni2015ecg}. Tereshchenko and Josephson reviewed studies on ECG frequencies and noted that ``the full spectrum of frequencies producing the QRS complex has not been adequately explored''~\cite{tereshchenko2015frequency}.\par 
Since there are no standard frequency separation methods for ECG signals~\cite{tereshchenko2015frequency}, we extract the logarithm of the average energy of five frequency bands (1-- 4 Hz, 4 -- 8 Hz, 8 -- 14 Hz, 14 -- 31 Hz, and 31 -- 50 Hz) from two ECG channels of the DREAMER dataset. As a result, we extract 10-dimensional features from the ECG signals.
\subsubsection{Eye movement features}
The eye movement data in the SEED dataset recorded using SMI ETG eye-tracking glasses$^{5}$ provide various types of parameters such as pupil diameters, fixation positions and durations, saccade information, blink details, and other event statistics. Although emotional changes cause fluctuations in pupil diameter, environmental luminance is the main reason for pupil diameter changes. Consequently, we adopt a principal component analysis-based method to remove the changes caused by lighting conditions~\cite{soleymani2012multimodal}.\par
The eye movement signals acquired by SMI ETG eye-tracking glasses contain both statistical features, such as blink information, and computational features such as temporal and frequency features. Table \ref{table:eye_features} shows all 33 eye movement features
used in this paper. Therefore, the total number of dimensions of the eye movement features is 33.
\begin{table}[!t]
\caption{Summary of extracted eye movement features.}
\centering
\begin{tabular}{|l|l|}
\hline
 Eye movement parameters & Extracted features \\
 \hline
 \multirow{4}{*}{Pupil diameter (X and Y)} & Mean, standard deviation, \\
                                          & DE in four bands\\
                                          & (0--0.2Hz,0.2--0.4Hz,\\
                                          & 0.4--0.6Hz,0.6--1Hz)\\
 \hline
 Disperson (X and Y) & Mean, standard deviation \\
 \hline
 Fixation duration (ms) & Mean, standard deviation \\
 \hline
 Blink duration (ms) & Mean, standard deviation \\
 \hline
 \multirow{3}{*}{Saccade} & Mean and standard deviation of \\
                          & saccade duration(ms) and \\
                          & saccade amplitude($^\circ$)\\
\hline
\multirow{9}{*}{Event statistics} & Blink frequency,\\
                                  & fixation frequency,\\
                                  & fixation duration maximum,\\
                                  & fixation dispersion total,\\
                                  & fixation dispersion maximum,\\
                                  & saccade frequency,\\
                                  & saccade duration average,\\
                                  & saccade amplitude average,\\
                                  & saccade latency average.\\
\hline
\end{tabular}%
\label{table:eye_features}
\end{table}
\begin{table*}[ht]
\caption{Summary of the DCCA structures for five different datasets}
\centering
\begin{tabular}{|l|c|l|r|}
\hline
 \multicolumn{1}{|c|}{Datasets} & \#Hidden Layers & \multicolumn{1}{|c|}{\#Hidden Units} & \multicolumn{1}{c|}{Output Dimensions} \\
 \hline
SEED & 6 & 400$\pm$40, 200$\pm$20, 150$\pm$20, 120$\pm$10, 60$\pm$10, 20$\pm$2
& 20 \\ \hline
SEED-IV & 7 & 400$\pm$40, 200$\pm$20, 150$\pm$20, 120$\pm$10, 90$\pm$10,
60$\pm$10, 20$\pm$2 & 20 \\ \hline
SEED-V & 2 &  searching for the best numbers between 50 and 200 & 12 \\ \hline

DEAP & 7 & 1500$\pm$50, 750$\pm$50, 500$\pm$25, 375$\pm$25, 130$\pm$20,
65$\pm$20, 30$\pm$20 & 20 \\ \hline
DREAMER & 2 & searching for the best numbers between 10 and 200 & 5 \\ \hline 
\end{tabular}%
\label{table:model_init}
\end{table*}
\subsubsection{Peripheral physiological signal features}
For peripheral physiological signals from the DEAP dataset, we calculate statistical features in the temporal domain, including the maximum value, minimum value, mean value, standard deviation, variance, and squared sum. Since there are 8 channels for the peripheral physiological signals, we extract 48 ($6\times8$)-dimensional features.
\subsection{Model training}
For the SEED dataset, the DE features of the first 9 movie clips are used as training data, and those of the remaining 6 movie clips are used as test data. 
In this paper, we build subject-dependent models to classify three types of emotions (happy, sad, and neutral), which is the same as in our previous work~\cite{lu2015combining,liu2016emotion,tang2017multimodal}.\par
A similar training-testing separation scheme is applied to the SEED-IV dataset. There are 24 trials for each session, and we use the data from the first 16 trials as the training data and the data from the remaining 8 trials as the test data~\cite{zheng2018emotionmeter}. DCCA is trained to recognize four emotions (happy, sad, fear, and neutral)\par 
For the SEED-V dataset, the training-testing separation strategy is the same as that used by Zhao {\it et .al}~\cite{zhao2019classification}. We adopt three-fold cross-validation to evaluate the performance of DCCA on five emotion (happy, sad, fear, neutral, and disgust) recognition tasks. Since the participant watched 15 movie clips (the first 5 clips, the middle 5 clips and the last 5 clips) and participated in three sessions, we concatenate features of the first 5 clips from three sessions ({\it i.e.}, we concatenate features extracted from 15 movie clips) as the training data for fold one (with a similar operation for folds two and three). 
\par 
For the DEAP dataset, we build a subject-dependent model with a 10-fold cross-validation on two binary classification tasks and a four-emotion
recognition task:
\begin{itemize}
\item Binary classifications: arousal-level and valence-level classification
with a threshold of 5.
\item Four-category classification: high arousal, high valence (HAHV); high
arousal, low valence (HALV); low arousal, high valence (LAHV); and low arousal,  low valence (LALV).
\end{itemize}
\par 
For the DREAMER dataset, we utilize leave-one-out cross-validation ({\it i.e.},  18-fold validation) to evaluate the performance of DCCA on three binary classification tasks (arousal, valence, and dominance), which is the same as that used by Song {\it et al.}~\cite{song2018eeg}.\par 
For these five different datasets, DCCA uses different hidden layers, hidden units, and output dimensions. Table \ref{table:model_init} summarizes the DCCA structures for these datasets. For all five datasets, the learning rate, batch size, and regulation parameter of DCCA are set to 0.001, 100, and $1e^{-8}$, respectively.
\par 
\section{Experimental results}
\label{sec:res}
\subsection{SEED, SEED-IV, and DEAP Datasets}
In this section, we summarize our previous results on SEED, SEED-IV, and DEAP datasets~\cite{qiu2018multi}.
Table \ref{table:res_seed} lists the results obtained by seven existing methods and DCCA on the SEED dataset. 
Lu and colleagues applied concatenation fusion, MAX fusion and fuzzy integral to fuse multiple modalities and demonstrated that the fuzzy integral fusion method achieved the accuracy of 87.59\%~\cite{lu2015combining}. Liu {\it et al.}~\cite{liu2016emotion} and Tang {\it et al.}~\cite{tang2017multimodal} improved multimodal methods, obtaining accuracies of 91.01\% and 94.58\%, respectively. Recently, Yang and colleagues~\cite{yang2018eeg} build a single-layer feedforward network (SLFN) with subnetwork nodes and achieved an accuracy of 91.51\%. Song and colleagues~\cite{song2018eeg} proposed DGCNN and obtained a classification accuracy of 90.40\%. As seen from Table \ref{table:res_seed}, DCCA achieves the best result of 94.58\% among the eight different methods.\par 
\begin{table}[h!]
\caption{The mean accuracy rates (\%) and standard deviations (\%) of seven existing methods and DCCA on the SEED dataset}
\centering
\begin{tabular}{l|l|r}
\hline
 Methods & Mean & \multicolumn{1}{c}{Std} \\
 \hline
Concatenation~\cite{lu2015combining} & 83.70 & - \\ \hline
MAX~\cite{lu2015combining} & 81.71 & - \\ \hline 
FuzzyIntegral~\cite{lu2015combining} & 87.59 & 19.87 \\ \hline 
BDAE~\cite{liu2016emotion} & 91.01 & 8.91 \\ \hline
DGCNN~\cite{song2018eeg} & 90.40 & 8.49 \\ \hline  
SLFN with subnetwork nodes~\cite{yang2018eeg} & 91.51 & --\\ \hline 
Bimodal-LSTM~\cite{tang2017multimodal} & 93.97 & 7.03 \\ \hline  
DCCA & \textbf{94.58} & \textbf{6.16} \\ \hline 
\end{tabular}%
\label{table:res_seed}
\end{table}
Table \ref{table:res_seed_iv} gives the results of five different methods on the SEED-IV dataset. We can observe from Table \ref{table:res_seed_iv} that for the SVM classifier, the four emotion states are recognized with a 75.88\% mean accuracy rate, and the BDAE model improved the result to 85.11\%. DCCA outperforms the aforementioned two methods, with an 87.45\% mean accuracy rate.
\par
\begin{table}[h!]
\caption{The mean accuracy rates (\%) and standard deviations (\%) of four existing methods and DCCA on the SEED-IV dataset}
\centering
\begin{tabular}{l|l|r}
\hline
 Methods & Mean & \multicolumn{1}{c}{Std} \\
 \hline
Concatenation & 77.63 & 16.43 \\ \hline
MAX & 68.99 & 17.14 \\ \hline
FuzzyIntegral & 73.55 & 16.72 \\ \hline 
BDAE~\cite{zheng2018emotionmeter} & 85.11 & 11.79 \\ \hline  
DCCA & \textbf{87.45} & \textbf{9.23} \\ \hline 
\end{tabular}%
\label{table:res_seed_iv}
\end{table}
Two classification schemes are adopted on the DEAP dataset. Table \ref{table:res_deap} shows the results of binary classifications. As we can observe, DCCA achieves the best results in both arousal classification (84.33\%) and valence classification (85.62\%) tasks. \par
For the four-category classification task on the DEAP dataset, Zheng and colleagues~\cite{zheng2017identifying} adopted the GELM model and achieved an accuracy of 69.67\%. Chen {\it et al.}~\cite{Chen2017A} proposed a three-stage decision framework that outperformed KNN and SVM with an accuracy rate of 70.04\%. The DCCA model achieved a mean accuracy rate of  88.51\%, which is more than 18\% higher than the existing methods.
\begin{table}[h!]
\caption{The mean accuracy rates (\%) and standard deviation (\%) of three existing methods and DCCA for the two binary emotion classification tasks on the DEAP dataset.}
\centering
\begin{tabular}{l|l|l}
\hline
 Methods & Arousal & Valence\\
 \hline
BDAE~\cite{liu2016emotion} & 80.50/3.39 & 85.20/4.47 \\ \hline 
MESAE~\cite{yin2017recognition} & 84.18/- & 83.04/- \\ \hline 
Bimodal-LSTM~\cite{tang2017multimodal} & 83.23/2.61 & 83.82/5.01 \\ \hline 
DCCA & \textbf{84.33/2.25} & \textbf{85.62/3.48}\\ \hline 
\end{tabular}%
\label{table:res_deap}
\end{table}
\begin{table}[h!]
\caption{The mean accuracy rates (\%) and standard deviations (\%) of two existing methods and DCCA for the four-category classification task on the DEAP dataset.}
\centering
\begin{tabular}{l|l}
\hline
 Methods & Acc  \\
 \hline
Three-stage decision Framework~\cite{Chen2017A} & 70.04/- \\ \hline 
GELM~\cite{zheng2017identifying} & 69.67/-  \\ \hline  
DCCA & \textbf{88.51/8.52} \\ \hline 
\end{tabular}%
\label{table:res_deap_four}
\end{table}
\par 
From the experimental results mentioned above, we can see that DCCA outperforms the existing methods on the SEED, SEED-IV, and DEAP datasets. 
\subsection{SEED-V Dataset}
We examine the effectiveness of DCCA on the SEED-V dataset, which contains
multimodal signals of five emotions (happy, sad, fear, neutral, and disgust).\par
\begin{figure}[h!]
\centering 
\includegraphics[width=.4\textwidth]{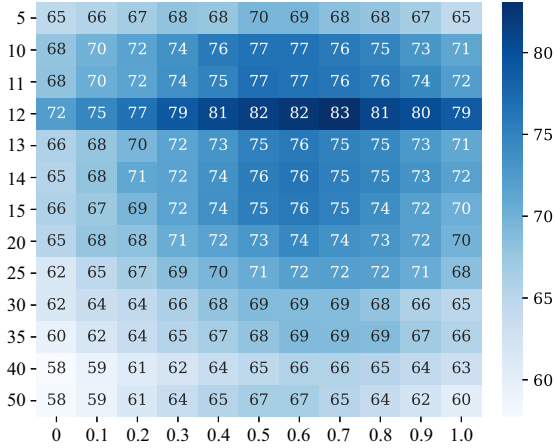}
\caption{Selection of the best output dimension and EEG weight of DCCA on the SEED-V dataset. Each row represents the number of output dimensions, and each column denotes the weight ($\alpha_1$) of the EEG features.}
\label{fig:grid_search}
\end{figure}

 
We perform a series of experiments to choose the best output dimension and  fusion coefficients ($\alpha_1$ and $\alpha_2$ in Eq. (\ref{eq:weighted_sum_fusion})) for DCCA. We adopt the grid search method with output dimensions ranging from 5 to 50 and coefficients for the EEG features ranging from 0 to 1, {\it{i.e.} $\alpha_1=[0,0.1,0.2,\cdots,0.9,1.0]$}. Since $\alpha_1+\alpha_2=1$, we can calculate the weight for the other modality via $\alpha_2=1-\alpha_1$.
Figure \ref{fig:grid_search} shows the heat map of the experimental results of the grid search. Each row in Fig. \ref{fig:grid_search} gives different output dimensions, and each column is the weight of the EEG features ($\alpha_1$). The numbers in blocks are the accuracy rates, which are rounded to integers for simplicity. According to Fig. \ref{fig:grid_search}, we set the output dimension to 12 and the weight of the EEG features to 0.7 ({\it i.e.}, $\alpha_1=0.7,\alpha_2=0.3$).\par 
\begin{figure}[h!]
\centering 
\includegraphics[width=.45\textwidth]{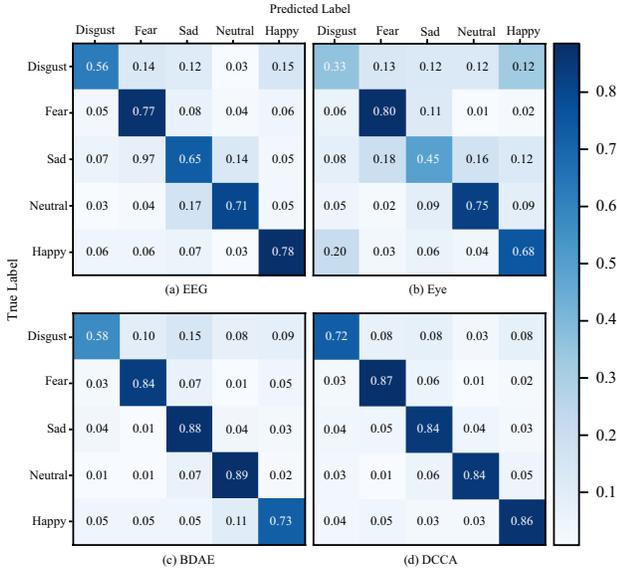}
\caption{Comparison of the confusion matrices of different methods on the SEED-V dataset. Subfigures (a), (b), and (c) are the confusion matrices from ~\cite{zhao2019classification} for SVM classifiers of unimodal features and BDAE model of multimodal features. Subfigure (d) is the confusion matrix of DCCA.}
\label{fig:seed_5_confusing_matrix}
\end{figure}
Table \ref{table:res_seed_v} summarizes the emotion recognition results on the SEED-V dataset.
Zhao and colleagues~\cite{zhao2019classification} adopted feature-level concatenation and the bimodal deep autoencoder (BDAE) for fusing multiple modalities, and achieved mean accuracy rates of 73.65\% and 79.70\%, respectively. In addition to feature-level concatenation, we also implement MAX fusion and fuzzy integral fusion strategies here. As shown in Table \ref{table:res_seed_v}, the MAX fusion and fuzzy integral fusion yielded mean accuracy rates of 73.14\% and 73.62\%, respectively. The mean accuracy rate of DCCA is 83.08\%, which is the best result among the five fusion strategies.\par 
\begin{table}[h]
\caption{The mean accuracy rates (\%) and standard deviations (\%) of four existing methods and DCCA on the SEED-V dataset}
\centering
\begin{tabular}{l|l|l}
\hline
 Methods & Mean & Std \\
 \hline 
Concatenation~\cite{zhao2019classification} & 73.65 & 8.90 \\ \hline 
MAX & 73.17 & 9.27 \\ \hline 
FuzzyIntegral & 73.24 & 8.72 \\ \hline  
BDAE~\cite{zhao2019classification} & 79.70 & \textbf{4.76} \\ \hline  
DCCA & \textbf{83.08} & 7.11 \\ \hline 
\end{tabular}%
\label{table:res_seed_v}
\end{table}
Figure \ref{fig:seed_5_confusing_matrix} depicts the confusion matrices of the DCCA
model and the models
adopted by Zhao and colleagues~\cite{zhao2019classification}.
Figures. \ref{fig:seed_5_confusing_matrix}(a),
(b) and (c) are the confusion matrices for the EEG features, eye movement features, and
the BDAE model,
respectively. Figure \ref{fig:seed_5_confusing_matrix}(d) depicts the
confusion matrix for the DCCA
model. 
From Figs. \ref{fig:seed_5_confusing_matrix}(a), (b), and (d), for each of
the five emotions, DCCA achieves a higher accuracy, indicating that 
emotions are better represented and more easily classified in the coordinated hyperspace $\mathcal{S}$ transformed by DCCA.\par 
From Figs. \ref{fig:seed_5_confusing_matrix}(a) and (c), compared with the unimodal results of the EEG features, the BDAE model achieves worse classification results on the happy emotion, suggesting that the BDAE model might not take full advantage of different modalities for the happy emotion. Comparing Figs. \ref{fig:seed_5_confusing_matrix}(c) and (d), 
DCCA largely improved the classification results on disgust and happy emotion recognition tasks compared with the BDAE model, implying that DCCA is more effective in fusing multiple modalities.
 \par 
\begin{figure*}[h!]
\centering
\includegraphics[width=.8\textwidth]{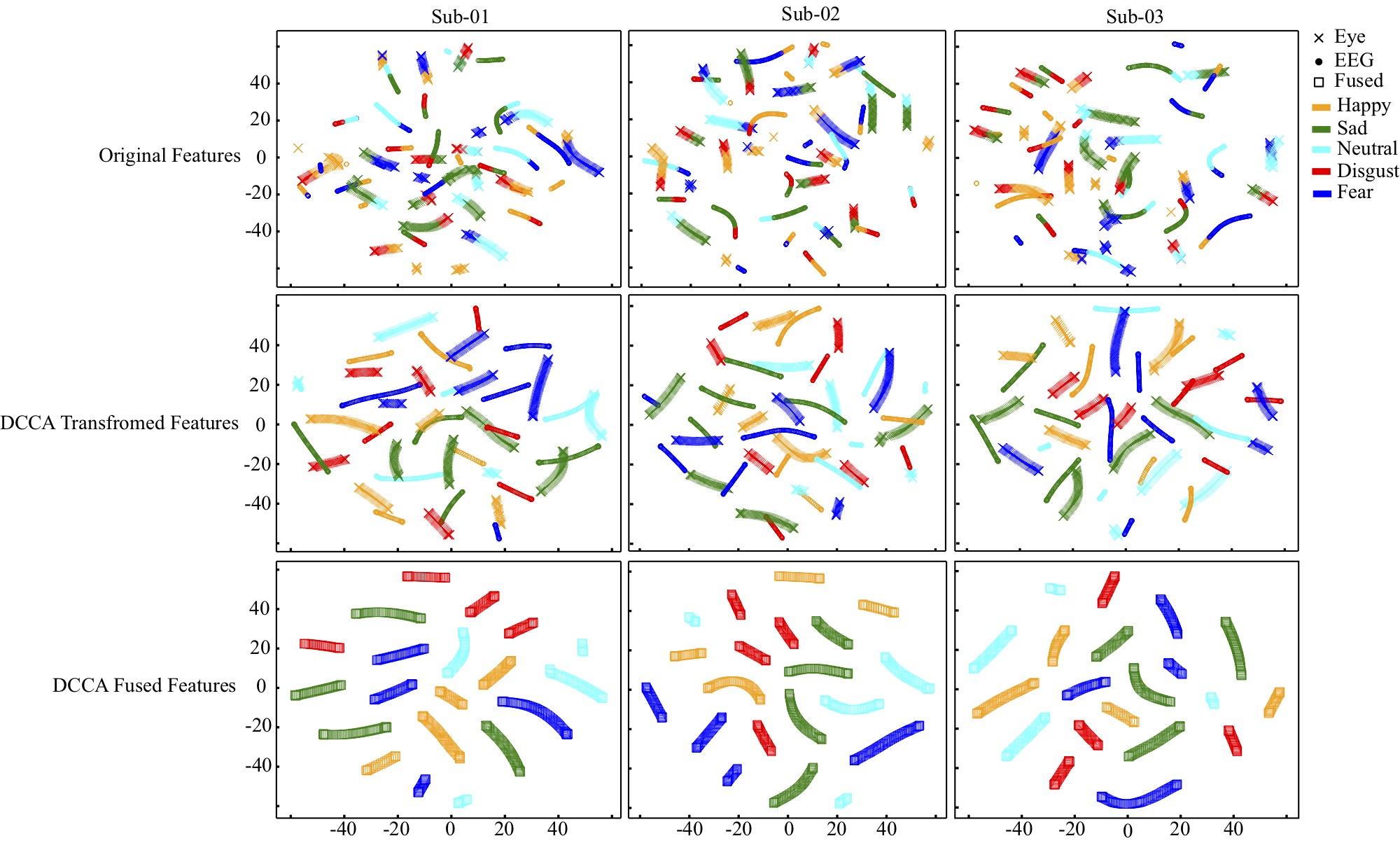}	
\caption{Feature distribution visualization by the t-SNE algorithm. The original features,
transformed features, and fused features from the three subjects are presented. The different colors stand for different emotions, and the different markers indicate different features.}
\label{fig:tsne}
\end{figure*}
\begin{figure*}[h!]
\centering
\includegraphics[width=.8\textwidth]{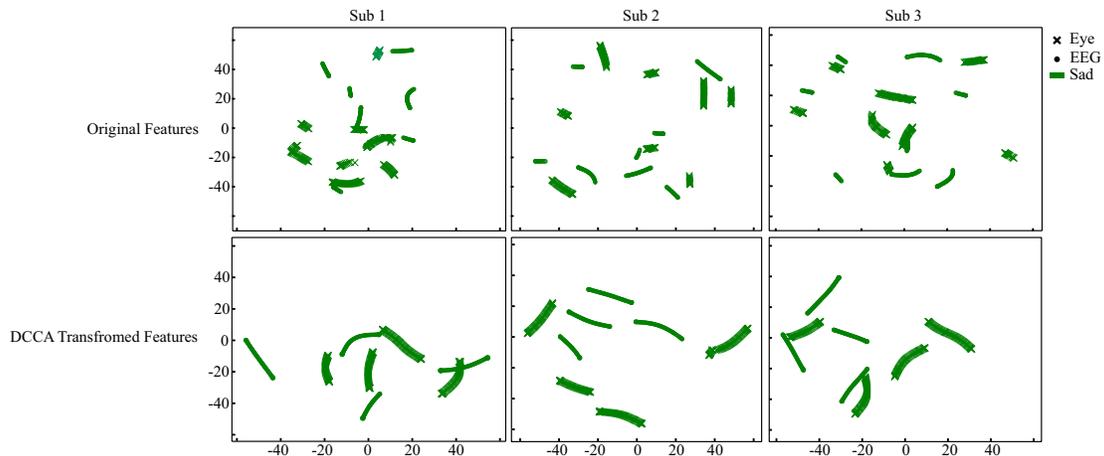}	
\caption{Distributions of EEG and eye movement features for the sad emotion. The transformed features have more compact distributions from both inter-modality and intra-modality perspectives.}
\label{fig:distribution}
\end{figure*}
To analyze the coordinated hyperspace $\mathcal{S}$ of DCCA,
we utilized the t-SNE algorithm to visualize the space of the  
original features and the coordinated hyperspace of the transformed features and fused features. 
Figure \ref{fig:tsne} presents 
a visualization of the features from three participants. The first row shows the original features, 
the second row depicts the transformed features, and the last row presents the 
fused features. The different colors stand for different emotions, and the 
different markers are different modalities.
We can make the following observations:
\begin{itemize}
	\item Different emotions are disentangled in the coordinated hyperspace $\mathcal{S}$. For original features, there are more overlaps among different emotions (different colors presenting substantial overlap), which lead to poorer emotional representation. After the DCCA transformation, different emotions become relatively independent, and the overlapping areas are considerably reduced. This indicates that the transformed features have improved emotional representation capabilities compared with the original features. Finally, after multimodal fusion, different emotions (`$\Box$' of different colors in the last row) are completely separated, and there is no overlapping area, indicating that the merged features also have good emotional representation ability.
	\item Different modalities have homogeneous distributions in the coordinated hyperspace $\mathcal{S}$. To make this observation more obvious, we separate and plot the distributions of the EEG and eye movement features under the sad emotion in Fig.~\ref{fig:distribution}. From the perspectives of both inter-modality and intra-modality distributions, the original EEG features (`$\circ$' marker) and eye movement features (`$\times$' marker) are separated from each other. After the DCCA transformation, the EEG features and the eye movement features have more compact distributions, indicating that the coordinated hyperspace preserves shared emotion-related information and discards irrelevant information.
\end{itemize}
\par 
Figures \ref{fig:tsne} and \ref{fig:distribution} qualitatively show that DCCA maps original EEG and eye movement features into a coordinated hyperspace $\mathcal{S}$ where emotions are better represented since only emotion related information is preserved.\par
\begin{figure*}[h]
\centering
\includegraphics[width=.8\textwidth]{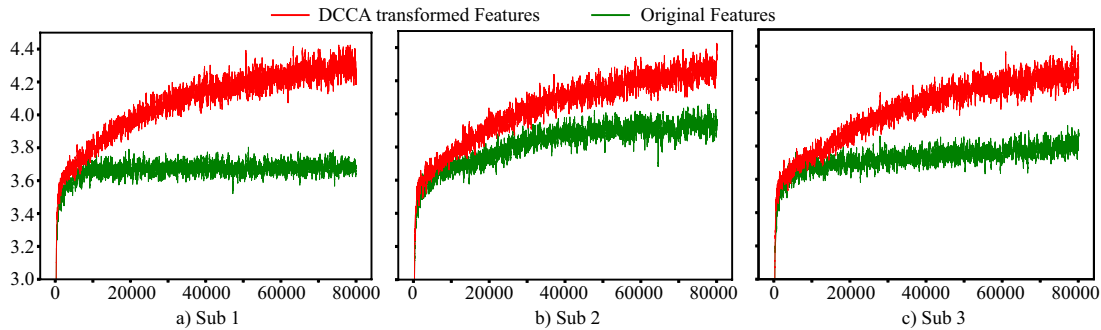}	
\caption{Mutual information (MI) estimation with MINE. The green curve shows the 
estimated MI for the original EEG features and eye movement
features. The red curve depicts the MI for the 
transformed features. The $x$ axis is the epoch number of the deep neural network used to estimate MI, and the $y$ axis is the estimated MI. Moving average smoothing is used to smooth the curves.}
\label{fig:mi}
\end{figure*}

 
Furthermore, we calculated the mutual information of the original features and transformed features to support our claims quantitatively.
Figure \ref{fig:mi} presents the mutual information of three participants estimated by MINE. The green curves depict the mutual information of the original EEG and eye movement features, and the red curves are the estimated mutual information of the transformed features. The transformed features have more mutual information than the original features, indicating that EEG and eye movement features in the coordinated hyperspace provide more shared emotion-related information, which is consistent with observations from Figs. \ref{fig:tsne} and \ref{fig:distribution}.\par 

\subsection{Robustness Analysis on the SEED-V Dataset}
EEG signals have a low signal-to-noise ratio (SNR) and are easily
interfered with by external environmental noise. To compare the noise robustness of DCCA with that of the existing methods, we designed two experimental schemes on noisy datasets: 1) we added Gaussian noise of different variances to both the EEG and eye movement features. To highlight the influence of noise, we added noise to the normalized features since the directly extracted features are much larger than the generated noise (which is mostly less than 1). 2) Under certain extreme conditions, EEG signals may be overwhelmed by noise. To simulate this situation, we randomly replace different proportions (10\%, 30\%, and 50\%) of EEG features with noise with a normal distribution ($X\sim\mathcal{N}(0,1)$), gamma distribution ($X\sim\Gamma(1,1)$), and uniform distribution ($X\sim\mathcal{U}[0,1]$). Specifically, for DCCA, we also examine the effect of different weight coefficients on the robustness of the model. In this paper, we compare the performance of three different combinations of coefficients, {\it{i.e.}}, $\alpha_1=0.3$ (DCCA-0.3), $\alpha_1=0.5$ (DCCA-0.5), and $\alpha_1=0.7$ (DCCA-0.7).\par 
\subsubsection{Adding Gaussian noise}
\begin{figure}[h!]
\centering
\includegraphics[width=.35\textwidth]{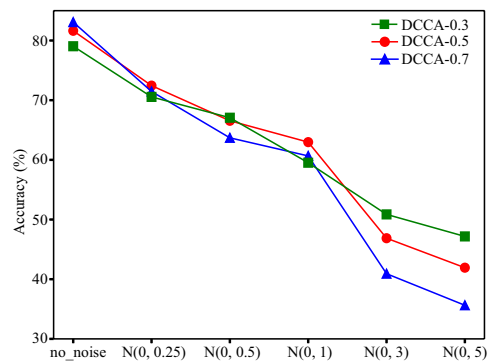}	
\caption{Performance of DCCA with different weight combinations when adding Gaussian noise of different variances. The robustness of DCCA decreases as the weight of the EEG features increases.}
\label{fig:dcca_within_adding}
\end{figure}

First, we investigate the robustness of different weight combinations in DCCA after adding Gaussian noise of different variances to both the EEG and eye movement features. Figure \ref{fig:dcca_within_adding} depicts the results. Although the model achieves the highest classification accuracy when the EEG weight is set to 0.7, it is also more susceptible to noise. The robustness of the model decreases as the weight of the EEG features increases. Since a larger EEG weight leads to more EEG components in the fused features, we might conclude that EEG features are more sensitive to noise than are eye movement features.\par
\begin{figure}[h!]
\centering
\includegraphics[width=.35\textwidth]{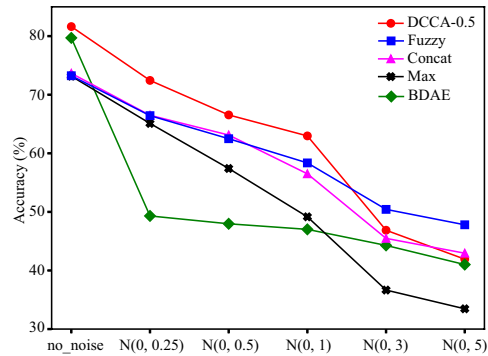}	
\caption{Model performances after adding Gaussian noise of different variances. The accuracies drops after noise is added to the original training data. DCCA obtains the best performance when the noise is less than $\mathcal{N}(0,1)$. When the noise is stronger than $\mathcal{N}(0,3)$, the fuzzy integral fusion strategy performed best.}
\label{fig:robust}
\end{figure}

 
Next, we compare the robustness of different models under Gaussian noise with different variances. Taking both classification performance and robustness into consideration, we use DCCA with an EEG weight set to 0.5. Figure \ref{fig:robust} shows the performances of the various models.
The performance decreases with increasing variances of the 
Gaussian noise. DCCA obtains the best performance when the noise is lower 
than or equal to $\mathcal{N}(0,1)$. The performance of the fuzzy integral 
fusion strategy exceeds DCCA when the noise is stronger than or equal to 
$\mathcal{N}(0,3)$. The BDAE model performs poorly under noisy conditions even when minimal noise is added to the training samples, the performance of the BDAE model is greatly reduced.
\par 
\begin{table*}[h!]
\renewcommand{\tabcolsep}{1.5mm}
\caption{Recognition results (Mean/Std (\%)) after replacing different proportions of EEG features with various types of noise. Five fusion strategies under various settings are compared, and the best results for each setting are in bold}
\centering
\begin{tabular}{l||r||r|r|r||r|r|r||r|r|r}
\hline
 \multirow{2}{*}{Methods} & \multirow{2}{*}{No noise} & \multicolumn{3}{c||}{Gaussian} & \multicolumn{3}{c||}{Gamma} & \multicolumn{3}{c}{Uniform} \\
 \cline{3-11}
 & & \multicolumn{1}{c|}{10\%} & \multicolumn{1}{c|}{30\%} & \multicolumn{1}{c||}{50\%} & \multicolumn{1}{c|}{10\%} & \multicolumn{1}{c|}{30\%} & \multicolumn{1}{c||}{50\%} & \multicolumn{1}{c|}{10\%} & \multicolumn{1}{c|}{30\%} & \multicolumn{1}{c}{50\%} \\
 \hline 
Concatenation & 73.65/8.90 & 70.08/8.79 & 63.13/9.05 & 58.32/7.51 & 69.71/8.51 & 62.93/8.46 & 57.97/8.14 & 71.24/10.56 & 66.46/9.38 & 61.82/8.35\\ \hline 
MAX & 73.17/9.27 & 67.67/8.38 & 58.29/8.41 &	 51.08/7.00 & 67.24/10.27 &	59.18/9.77 & 50.56/6.82 & 67.51/9.72 & 60.14/9.28 & 52.71/7.84\\ \hline 
FuzzyIntegral & 73.24/8.72 & 69.42/8.92 & 62.98/7.52 & 57.69/8.70 & 69.35/8.70 & 62.64/8.90 & 57.56/7.19 & 69.16/8.16 & 64.86/9.37 & 60.47/8.32 \\ \hline 
BDAE & 79.70/\textbf{4.76} & 47.82/7.77 & 45.89/7.82 & 44.51/7.43 & 45.27/\textbf{6.68} & 45.75/7.91 & 45.09/8.37 & 46.13/8.17 & 46.88/7.14 & 45.50/9.59 \\ \hline 
DCCA-0.3 & 79.04/7.32 & 76.57/7.63 & \textbf{73.00}/7.36 & \textbf{69.56}/7.02 & 76.87/7.99 & \textbf{73.06}/7.00 & \textbf{70.03}/7.17 & 75.69/\textbf{6.34} & \textbf{73.22}/6.50 & \textbf{70.01}/6.66 \\ \hline 
DCCA-0.5 & 81.62/6.95 & \textbf{77.92/6.63} & 71.77/6.55 & 65.21/6.24 & \textbf{78.29}/7.38 & 72.45/6.14 & 65.75/6.08 & \textbf{78.28}/7.16 & 73.20/6.96 & 68.01/7.08 \\ \hline 
DCCA-0.7 & \textbf{83.08}/7.11 & 76.27/7.02 & 68.48/\textbf{5.54} & 57.63/\textbf{5.15} & 76.82/7.01 & 68.54/\textbf{6.02} & 58.58/\textbf{5.44} & 77.39/8.43 & 69.80/\textbf{5.63} & 61.58/\textbf{5.38} \\ \hline
\end{tabular}%
\label{table:missing}
\end{table*}
\subsubsection{Replacing EEG features with noise}
Table \ref{table:missing} shows the detailed emotion recognition accuracies and standard deviations after replacing 10\%, 30\%, and 50\% percent of the EEG features with different noise distributions. The recognition accuracies decrease with increasing noise proportions. In addition, the performances of seven different settings under different noise distributions are very similar, indicating that noise distributions have limited influences on the recognition accuracies.
\par
\begin{figure}[ht!]
\centering
\includegraphics[width=.35\textwidth]{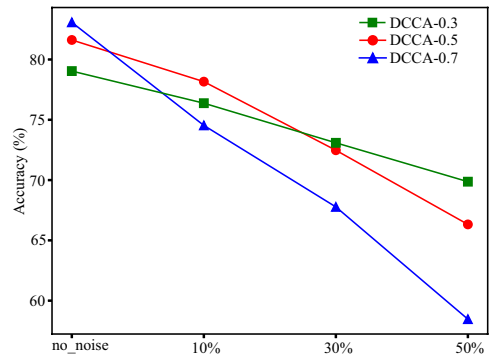}	
\caption{Performance of DCCA with different weight combinations after replacing the EEG features with noise.}
\label{fig:dcca_within_replacing}
\end{figure}

 
To better observe the changing tendency, we plot the average recognition accuracies under different noise distributions with the same noise ratio. Figure \ref{fig:dcca_within_replacing} shows the average accuracies for DCCA with different EEG weights. It is obvious that the performances decrease with increasing noise percentages and that the model robustness is inversely proportional to the ratio of the EEG modality. This is the expected performance. Since we only randomly replace EEG features with noise, larger EEG weights will introduce more noises to the fused features, resulting in a decrease in model robustness.\par
\begin{figure}[h!]
\centering
\includegraphics[width=.35\textwidth]{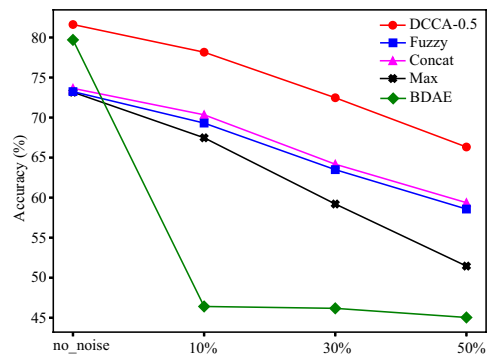}	
\caption{The trends of the average recognition accuracies of different noise distributions under the same noise ratio. The $x$-axis is the noise replacement ratio, and the $y$-axis stands for the mean accuracies.}
\label{fig:missing}
\end{figure}

 
Similar to Fig. \ref{fig:dcca_within_adding}, we also take DCCA-0.5, as a compromise between performance and robustness to compare with other multimodal fusion methods.
Figure \ref{fig:missing} depicts the trends of the accuracies of several models. It is obvious that DCCA performs the best, the concatenation fusion achieves a slightly better performance than the fuzzy integral fusion method, and the BDAE model again presents the worst performance.\par 
\par 
Combining Figs.~\ref{fig:robust} and \ref{fig:missing}, DCCA obtains the best performance under most noisy situations, whereas the BDAE model performed the worst under noisy conditions. This might be caused by the following:
\begin{itemize}
	\item As already discussed in previous sections, DCCA attemps to preserve emotion-related information and discard irrelevant information. This property prevents the model performance from rapidly deteriorating by neglecting negative information introduced by noise.
	\item The BDAE model minimizes the mean squared error which is sensitive to outliers~\cite{kim2012robust}. The noisy training features will cause the weights to deviate from the normal range, resulting in a rapid decline in model performance.
\end{itemize}
\subsection{DREAMER Dataset}
\begin{figure}[h]
\centering 
\includegraphics[width=.5\textwidth]{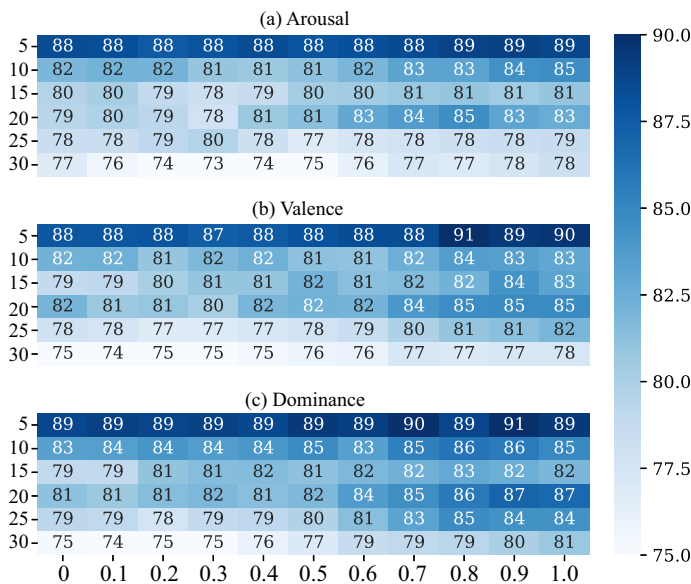}
\caption{Selecting the best output dimension and weight combinations of DCCA on the DREAMER dataset. The $X$-axis represents the weight for the EEG features, and the $Y$-axis represents the output dimensions.}
\label{fig:dreamer_grid_search}
\end{figure}

For DCCA, we choose the best output dimensions and weight combinations with a grid search. We select the output dimension from the set $[5, 10, 15, 20, 25, 30]$ and the EEG weight $\alpha_1$ in $[0, 0.1, \cdots, 0.9,1.0]$ for three binary classification tasks. Figures \ref{fig:dreamer_grid_search}(a), (b), and (c) depict the heat maps of the grid search for arousal, valence, and dominance classifications, respectively. According to Fig. \ref{fig:dreamer_grid_search}, we choose $\alpha_1=0.9$ and $\alpha_2=0.1$ for the arousal classification, $\alpha_1=0.8$ and $\alpha_2=0.2$ for the valence classification, and $\alpha_1=0.9$ and $\alpha_2=0.1$ for the dominance classification.
\par 
For BDAE, we select the best output dimensions from $[700, 500, 200,
170, 150, 130, 110, 90, 70, 50]$,
and leave-one-out cross-validation is used to evaluate the BDAE model.\par 
Table \ref{table:res_dreamer} gives comparison results of the different methods. 
Katsigiannis and Ramzan released this dataset, and they achieved accuracy rates of 62.32\%, 61.84\%, and 61.84\% on arousal, valence and dominance classification tasks, respectively~\cite{katsigiannis2017dreamer}.
Song and colleagues conducted a series of experiments on this dataset with SVM, graphSLDA, GSCCA, and DGCNN. DGCNN achieved accuracy rates of 85.54\% for arousal classification, 86.23\% for valence classification, and 85.02\% for dominance classification. From Table \ref{table:res_dreamer}, we can see that BDAE and DCCA adopted in this paper outperform DGCNN. For BDAE, the recognition results for arousal, valence, and dominance are 88.57\%, 86.64\%, and 89.52\%, respectively. DCCA achieves the best performance among all seven methods: 88.99\%, 90.57\%, and 90.67\% for arousal, valence, and dominance level recognitions, respectively. \par 
\begin{table}[h]
\caption{Comparison of performances (mean/Std, \%) on the DREAMER dataset. Three binary classification tasks are evaluated: arousal-level, valence-level, and dominance-level classifications}
\centering
\begin{tabular}{l|r|r|r}
\hline
 Methods & \multicolumn{1}{c|}{Arousal} & \multicolumn{1}{c|}{Valence} & \multicolumn{1}{c}{Dominance} \\
 \hline
Fusion EEG \& ECG~\cite{katsigiannis2017dreamer} & 62.32/- & 61.84/- & 61.84/- \\ \hline 
SVM~\cite{song2018eeg} & 68.84/24.92 & 60.14/33.34 & 75.84/20.76 \\ \hline
GraphSLDA~\cite{song2018eeg} & 68.12/17.53 & 57.70/13.89 & 73.90/15.85  \\ \hline
GSCCA~\cite{song2018eeg} & 70.30/18.66 & 56.65/21.50 & 77.31/15.44 \\ \hline
DGCNN~\cite{song2018eeg} & 84.54/10.18 & 86.23/12.29 & 85.02/10.25 \\ \hline   
BDAE & 88.57/4.40 & 86.64/7.48 & 89.52/6.18 \\ \hline 
Our method & \textbf{88.99/2.84} & \textbf{90.57/4.11} & \textbf{90.67/4.33}
\\ \hline
\end{tabular}%
\label{table:res_dreamer}
\end{table}

\section{Conclusion}
\label{sec:conclusion}
In this paper, we have introduced deep canonical correlation analysis (DCCA) to multimodal emotion recognition.
We have systematically evaluated the performance of DCCA on five multimodal emotion datasets (the SEED, SEED-IV, SEED-V, DEAP and DREAMER datasets) and compared DCCA with the existing emotion recognition methods. 
Our experimental results demonstrate that DCCA is superior to the existing methods for multimodal emotion recognition.\par  
We have analyzed properties of the transformed features in the coordinated hyperspace $\mathcal{S}$. By applying t-SNE method, we have found qualitatively that: 1) different emotions are better represented since they are disentangled in the coordinated hyperspace; and 2) different modalities have compact distributions from both inter-modality and intra-modality perspectives. We have applied mutual information neural estimation (MINE) algorithm to compare the mutual information of original features and transformed features quantitatively. The experimental results show that the features transformed by DCCA have higher mutual information, indicating that DCCA transformation processes preserve emotion-related information and discard irrelevant information.\par 
We have investigated the robustness of DCCA on noised datasets under two schemes. By adding Gaussian noise of different variances to both EEG and eye movement features, we have demonstrated that DCCA performs best when the noise is smaller than or equal to $N(0,1)$. After replacing 10\%, 30\%, and 50\% percentage of EEG features with normal distribution, gamma distribution, and uniform distribution, we have shown that DCCA has the best performance for multimodal emotion recognition.

%


\ifCLASSOPTIONcaptionsoff
  \newpage
\fi



\bibliographystyle{IEEEtran}
\bibliography{2019Journal}

\begin{thebibliography}{10}
\providecommand{\url}[1]{#1}
\csname url@samestyle\endcsname
\providecommand{\newblock}{\relax}
\providecommand{\bibinfo}[2]{#2}
\providecommand{\BIBentrySTDinterwordspacing}{\spaceskip=0pt\relax}
\providecommand{\BIBentryALTinterwordstretchfactor}{4}
\providecommand{\BIBentryALTinterwordspacing}{\spaceskip=\fontdimen2\font plus
\BIBentryALTinterwordstretchfactor\fontdimen3\font minus
  \fontdimen4\font\relax}
\providecommand{\BIBforeignlanguage}[2]{{%
\expandafter\ifx\csname l@#1\endcsname\relax
\typeout{** WARNING: IEEEtran.bst: No hyphenation pattern has been}%
\typeout{** loaded for the language `#1'. Using the pattern for}%
\typeout{** the default language instead.}%
\else
\language=\csname l@#1\endcsname
\fi
#2}}
\providecommand{\BIBdecl}{\relax}
\BIBdecl

\bibitem{picard2000affective}
R.~W. Picard, \emph{Affective computing}.\hskip 1em plus 0.5em minus
  0.4em\relax MIT press, 2000.

\bibitem{picard2001toward}
R.~W. Picard, E.~Vyzas, and J.~Healey, ``Toward machine emotional intelligence:
  Analysis of affective physiological state,'' \emph{IEEE Transactions on
  Pattern Analysis \& Machine Intelligence}, no.~10, pp. 1175--1191, 2001.

\bibitem{el2011survey}
M.~El~Ayadi, M.~S. Kamel, and F.~Karray, ``Survey on speech emotion
  recognition: Features, classification schemes, and databases,'' \emph{Pattern
  Recognition}, vol.~44, no.~3, pp. 572--587, 2011.

\bibitem{ko2018brief}
B.~Ko, ``A brief review of facial emotion recognition based on visual
  information,'' \emph{Sensors}, vol.~18, no.~2, p. 401, 2018.

\bibitem{yadollahi2017current}
A.~Yadollahi, A.~G. Shahraki, and O.~R. Zaiane, ``Current state of text
  sentiment analysis from opinion to emotion mining,'' \emph{ACM Computing
  Surveys (CSUR)}, vol.~50, no.~2, p.~25, 2017.

\bibitem{zheng2015investigating}
W.-L. Zheng and B.-L. Lu, ``Investigating critical frequency bands and channels
  for {EEG}-based emotion recognition with deep neural networks,'' \emph{IEEE
  Transactions on Autonomous Mental Development}, vol.~7, no.~3, pp. 162--175,
  2015.

\bibitem{zheng2017identifying}
W.-L. Zheng, J.-Y. Zhu, and B.-L. Lu, ``Identifying stable patterns over time
  for emotion recognition from {EEG},'' \emph{IEEE Transactions on Affective
  Computing}, doi: 10.1109/TAFFC.2017.2712143.

\bibitem{yang2018eeg}
Y.~Yang, Q.~J. Wu, W.-L. Zheng, and B.-L. Lu, ``{EEG}-based emotion recognition
  using hierarchical network with subnetwork nodes,'' \emph{IEEE Transactions
  on Cognitive and Developmental Systems}, vol.~10, no.~2, pp. 408--419, 2018.

\bibitem{yin2017cross}
Z.~Yin, Y.~Wang, L.~Liu, W.~Zhang, and J.~Zhang, ``Cross-subject {EEG} feature
  selection for emotion recognition using transfer recursive feature
  elimination,'' \emph{Frontiers in Neurorobotics}, vol.~11, p.~19, 2017.

\bibitem{lin2011generalizations}
Y.-P. Lin, J.-H. Chen, J.-R. Duann, C.-T. Lin, and T.-P. Jung,
  ``Generalizations of the subject-independent feature set for music-induced
  emotion recognition,'' in \emph{2011 Annual International Conference of the
  IEEE Engineering in Medicine and Biology Society}.\hskip 1em plus 0.5em minus
  0.4em\relax IEEE, 2011, pp. 6092--6095.

\bibitem{wang2014emotional}
X.-W. Wang, D.~Nie, and B.-L. Lu, ``Emotional state classification from eeg
  data using machine learning approach,'' \emph{Neurocomputing}, vol. 129, pp.
  94--106, 2014.

\bibitem{kim2008}
J.~Kim and E.~Andr\'e, ``{Emotion recognition based on physiological changes in
  music listening},'' \emph{IEEE Transactions on Pattern Analysis and Machine
  Intelligence}, vol.~30, pp. 2067--2083, 2008.

\bibitem{vo2008coupling}
M.~L.-H. V{\~o}, A.~M. Jacobs, L.~Kuchinke, M.~Hofmann, M.~Conrad, A.~Schacht,
  and F.~Hutzler, ``The coupling of emotion and cognition in the eye:
  Introducing the pupil old/new effect,'' \emph{Psychophysiology}, vol.~45,
  no.~1, pp. 130--140, 2008.

\bibitem{poria2017}
S.~Poria, E.~Cambria, R.~Bajpai, and A.~Hussain, ``{A review of affective
  computing: from unimodal analysis to multimodal fusion},'' \emph{Information
  Fusion}, vol.~37, pp. 98--125, 2017.

\bibitem{zheng2018emotionmeter}
W.-L. {Zheng}, W.~{Liu}, Y.-F. {Lu}, B.-L. {Lu}, and A.~{Cichocki},
  ``Emotionmeter: A multimodal framework for recognizing human emotions,''
  \emph{IEEE Transactions on Cybernetics}, vol.~49, no.~3, pp. 1110--1122,
  March 2019.

\bibitem{soleymani2012multimodal}
M.~Soleymani, M.~Pantic, and T.~Pun, ``Multimodal emotion recognition in
  response to videos,'' \emph{IEEE Transactions on Affective Computing},
  vol.~3, no.~2, pp. 211--223, 2012.

\bibitem{6095505}
M.~{Soleymani}, M.~{Pantic}, and T.~{Pun}, ``Multimodal emotion recognition in
  response to videos,'' \emph{IEEE Transactions on Affective Computing},
  vol.~3, no.~2, pp. 211--223, April 2012.

\bibitem{8013713}
A.~{Mollahosseini}, B.~{Hasani}, and M.~H. {Mahoor}, ``Affectnet: A database
  for facial expression, valence, and arousal computing in the wild,''
  \emph{IEEE Transactions on Affective Computing}, vol.~10, no.~1, pp. 18--31,
  Jan 2019.

\bibitem{7112127}
M.~{Soleymani}, S.~{Asghari-Esfeden}, Y.~{Fu}, and M.~{Pantic}, ``Analysis of
  eeg signals and facial expressions for continuous emotion detection,''
  \emph{IEEE Transactions on Affective Computing}, vol.~7, no.~1, pp. 17--28,
  Jan 2016.

\bibitem{lu2015combining}
Y.-F. Lu, W.-L. Zheng, B.-B. Li, and B.-L. Lu, ``Combining eye movements and
  {EEG} to enhance emotion recognition,'' in \emph{Twenty-Fourth International
  Joint Conference on Artificial Intelligence}, 2015.

\bibitem{koelstra2012deap}
S.~Koelstra, C.~Muhl, M.~Soleymani, J.-S. Lee, A.~Yazdani, T.~Ebrahimi, T.~Pun,
  A.~Nijholt, and I.~Patras, ``{DEAP}: A database for emotion analysis; using
  physiological signals,'' \emph{IEEE Transactions on Affective Computing},
  vol.~3, no.~1, pp. 18--31, 2012.

\bibitem{sun2016combining}
B.~Sun, L.~Li, X.~Wu, T.~Zuo, Y.~Chen, G.~Zhou, J.~He, and X.~Zhu, ``Combining
  feature-level and decision-level fusion in a hierarchical classifier for
  emotion recognition in the wild,'' \emph{Journal on Multimodal User
  Interfaces}, vol.~10, no.~2, pp. 125--137, 2016.

\bibitem{Baltru2017Multimodal}
T.~{Baltrušaitis}, C.~{Ahuja}, and L.~{Morency}, ``Multimodal machine
  learning: A survey and taxonomy,'' \emph{IEEE Transactions on Pattern
  Analysis \& Machine Intelligence}, vol.~41, no.~2, pp. 423--443, 2017.

\bibitem{liu2016emotion}
W.~Liu, W.-L. Zheng, and B.-L. Lu, ``Emotion recognition using multimodal deep
  learning,'' in \emph{International Conference on Neural Information
  Processing}.\hskip 1em plus 0.5em minus 0.4em\relax Springer, 2016, pp.
  521--529.

\bibitem{tang2017multimodal}
H.~Tang, W.~Liu, W.-L. Zheng, and B.-L. Lu, ``Multimodal emotion recognition
  using deep neural networks,'' in \emph{International Conference on Neural
  Information Processing}.\hskip 1em plus 0.5em minus 0.4em\relax Springer,
  2017, pp. 811--819.

\bibitem{li2016emotion}
X.~Li, D.~Song, P.~Zhang, G.~Yu, Y.~Hou, and B.~Hu, ``Emotion recognition from
  multi-channel {EEG} data through convolutional recurrent neural network,'' in
  \emph{2016 IEEE International Conference on Bioinformatics and Biomedicine
  (BIBM)}.\hskip 1em plus 0.5em minus 0.4em\relax IEEE, 2016, pp. 352--359.

\bibitem{yin2017recognition}
Z.~Yin, M.~Zhao, Y.~Wang, J.~Yang, and J.~Zhang, ``Recognition of emotions
  using multimodal physiological signals and an ensemble deep learning model,''
  \emph{Computer Methods and Programs in Biomedicine}, vol. 140, pp. 93--110,
  2017.

\bibitem{andrew2013deep}
G.~Andrew, R.~Arora, J.~Bilmes, and K.~Livescu, ``Deep canonical correlation
  analysis,'' in \emph{International Conference on Machine Learning}, 2013, pp.
  1247--1255.

\bibitem{qiu2018multi}
J.-L. Qiu, W.~Liu, and B.-L. Lu, ``Multi-view emotion recognition using deep
  canonical correlation analysis,'' in \emph{International Conference on Neural
  Information Processing}.\hskip 1em plus 0.5em minus 0.4em\relax Springer,
  2018, pp. 221--231.

\bibitem{lahat2015multimodal}
D.~Lahat, T.~Adali, and C.~Jutten, ``Multimodal data fusion: an overview of
  methods, challenges, and prospects,'' \emph{Proceedings of the IEEE}, vol.
  103, no.~9, pp. 1449--1477, 2015.

\bibitem{D2015A}
S.~K. D'Mello and J.~Kory, ``A review and meta-analysis of multimodal affect
  detection systems,'' \emph{{ACM} Computing Surveys}, vol.~47, no.~3, pp.
  1--36, 2015.

\bibitem{hazarika2018self}
D.~Hazarika, S.~Gorantla, S.~Poria, and R.~Zimmermann, ``Self-attentive
  feature-level fusion for multimodal emotion detection,'' in \emph{2018 IEEE
  Conference on Multimedia Information Processing and Retrieval (MIPR)}.\hskip
  1em plus 0.5em minus 0.4em\relax IEEE, 2018, pp. 196--201.

\bibitem{ngiam2011multimodal}
J.~Ngiam, A.~Khosla, M.~Kim, J.~Nam, H.~Lee, and A.~Y. Ng, ``Multimodal deep
  learning,'' in \emph{International Conference on Machine Learning}, 2011, pp.
  689--696.

\bibitem{Monkaresi2012Classification}
H.~Monkaresi, M.~Sazzad, and R.~A. Calvo, ``Classification of affects using
  head movement, skin color features and physiological signals,'' in \emph{IEEE
  International Conference on Systems}, 2012.

\bibitem{guo2019}
K.~Guo, R.~Chai, H.~Candra, Y.~Guo, R.~Song, H.~Nguyen, and S.~Su, ``{A hybrid
  fuzzy cognitive map/support vector machine approach for EEG-based emotion
  classification using compressed sensing},'' \emph{International Journal of
  Fuzzy Systems}, vol.~21, pp. 263--273, 2019.

\bibitem{naim2014unsupervised}
I.~Naim, Y.~C. Song, Q.~Liu, H.~Kautz, J.~Luo, and D.~Gildea, ``Unsupervised
  alignment of natural language instructions with video segments,'' in
  \emph{Twenty-Eighth AAAI Conference on Artificial Intelligence}.\hskip 1em
  plus 0.5em minus 0.4em\relax AAAI Press, 2014, pp. 1558--1564.

\bibitem{frome2013devise}
A.~Frome, G.~S. Corrado, J.~Shlens, S.~Bengio, J.~Dean, T.~Mikolov
  \emph{et~al.}, ``Devise: A deep visual-semantic embedding model,'' in
  \emph{Advances in Neural Information Processing Systems}, 2013, pp.
  2121--2129.

\bibitem{hotelling1992relations}
H.~Hotelling, ``Relations between two sets of variates,'' in
  \emph{Breakthroughs in Statistics}.\hskip 1em plus 0.5em minus 0.4em\relax
  Springer, 1992, pp. 162--190.

\bibitem{hardoon2004canonical}
D.~R. Hardoon, S.~Szedmak, and J.~Shawe-Taylor, ``Canonical correlation
  analysis: An overview with application to learning methods,'' \emph{Neural
  Computation}, vol.~16, no.~12, pp. 2639--2664, 2004.

\bibitem{lai2000kernel}
P.~L. Lai and C.~Fyfe, ``Kernel and nonlinear canonical correlation analysis,''
  \emph{International Journal of Neural Systems}, vol.~10, no.~05, pp.
  365--377, 2000.

\bibitem{Klami2008Probabilistic}
A.~Klami and S.~Kaski, ``Probabilistic approach to detecting dependencies
  between data sets,'' \emph{Neurocomputing}, vol.~72, no.~1, pp. 39--46, 2008.

\bibitem{klami2013bayesian}
A.~Klami, S.~Virtanen, and S.~Kaski, ``Bayesian canonical correlation
  analysis,'' \emph{Journal of Machine Learning Research}, vol.~14, no. Apr,
  pp. 965--1003, 2013.

\bibitem{kim2007tensor}
T.-K. Kim, S.-F. Wong, and R.~Cipolla, ``Tensor canonical correlation analysis
  for action classification,'' in \emph{2007 IEEE Conference on Computer Vision
  and Pattern Recognition}.\hskip 1em plus 0.5em minus 0.4em\relax IEEE, 2007,
  pp. 1--8.

\bibitem{hardoon2011sparse}
D.~R. Hardoon and J.~Shawe-Taylor, ``Sparse canonical correlation analysis,''
  \emph{Machine Learning}, vol.~83, no.~3, pp. 331--353, 2011.

\bibitem{rasiwasia2014cluster}
N.~Rasiwasia, D.~Mahajan, V.~Mahadevan, and G.~Aggarwal, ``Cluster canonical
  correlation analysis,'' in \emph{Artificial Intelligence and Statistics},
  2014, pp. 823--831.

\bibitem{Grabisch2000Application}
M.~Grabisch and M.~Roubens, ``Application of the choquet integral in
  multicriteria decision making,'' \emph{Fuzzy Measures \& Integrals}, pp.
  348--374, 2000.

\bibitem{li2012gender}
B.~Li, X.-C. Lian, and B.-L. Lu, ``Gender classification by combining clothing,
  hair and facial component classifiers,'' \emph{Neurocomputing}, vol.~76,
  no.~1, pp. 18--27, 2012.

\bibitem{Tanaka1991A}
K.~Tanaka and M.~Sugeno, ``A study on subjective evaluations of printed color
  images,'' \emph{International Journal of Approximate Reasoning}, vol.~5,
  no.~5, pp. 213--222, 1991.

\bibitem{scikit-learn}
F.~Pedregosa, G.~Varoquaux, A.~Gramfort, V.~Michel, B.~Thirion, O.~Grisel,
  M.~Blondel, P.~Prettenhofer, R.~Weiss, V.~Dubourg, J.~Vanderplas, A.~Passos,
  D.~Cournapeau, M.~Brucher, M.~Perrot, and E.~Duchesnay, ``Scikit-learn:
  Machine learning in {P}ython,'' \emph{Journal of Machine Learning Research},
  vol.~12, pp. 2825--2830, 2011.

\bibitem{belghazi2018mine}
M.~I. Belghazi, A.~Baratin, S.~Rajeswar, S.~Ozair, Y.~Bengio, A.~Courville, and
  R.~D. Hjelm, ``Mine: mutual information neural estimation,'' \emph{arXiv
  preprint arXiv:1801.04062}, 2018.

\bibitem{li2019classification}
T.-H. Li, W.~Liu, W.-L. Zheng, and B.-L. Lu, ``Classification of five emotions
  from eeg and eye movement signals: Discrimination ability and stability over
  time,'' in \emph{9th International IEEE/EMBS Conference on Neural Engineering
  (NER)}.\hskip 1em plus 0.5em minus 0.4em\relax IEEE, 2019, pp. 607--610.

\bibitem{katsigiannis2017dreamer}
S.~Katsigiannis and N.~Ramzan, ``{DREAMER}: A database for emotion recognition
  through eeg and ecg signals from wireless low-cost off-the-shelf devices,''
  \emph{IEEE Journal of Biomedical and Health Informatics}, vol.~22, no.~1, pp.
  98--107, 2017.

\bibitem{duan2013differential}
R.-N. Duan, J.-Y. Zhu, and B.-L. Lu, ``Differential entropy feature for
  {EEG}-based emotion classification,'' in \emph{2013 6th International
  IEEE/EMBS Conference on Neural Engineering (NER)}.\hskip 1em plus 0.5em minus
  0.4em\relax IEEE, 2013, pp. 81--84.

\bibitem{shi2013differential}
L.-C. Shi, Y.-Y. Jiao, and B.-L. Lu, ``Differential entropy feature for
  {EEG}-based vigilance estimation,'' in \emph{2013 35th Annual International
  Conference of the IEEE Engineering in Medicine and Biology Society
  (EMBC)}.\hskip 1em plus 0.5em minus 0.4em\relax IEEE, 2013, pp. 6627--6630.

\bibitem{shi2010off}
L.-C. Shi and B.-L. Lu, ``Off-line and on-line vigilance estimation based on
  linear dynamical system and manifold learning,'' in \emph{2010 Annual
  International Conference of the IEEE Engineering in Medicine and
  Biology}.\hskip 1em plus 0.5em minus 0.4em\relax IEEE, 2010, pp. 6587--6590.

\bibitem{8219396}
Y.~{Hsu}, J.~{Wang}, W.~{Chiang}, and C.~{Hung}, ``Automatic ecg-based emotion
  recognition in music listening,'' \emph{IEEE Transactions on Affective
  Computing}, pp. 1--16, 2018.

\bibitem{zhao2016wireless}
M.~Zhao, F.~Adib, and D.~Katabi, ``Emotion recognition using wireless
  signals,'' in \emph{Proceedings of the 22nd Annual International Conference
  on Mobile Computing and Networking}.\hskip 1em plus 0.5em minus 0.4em\relax
  ACM, 2016, pp. 95--108.

\bibitem{scher1960frequency}
A.~M. SCHER and A.~C. YOUNG, ``Frequency analysis of the electrocardiogram,''
  \emph{Circulation Research}, vol.~8, no.~2, pp. 344--346, 1960.

\bibitem{shufni2015ecg}
S.~A. Shufni and M.~Y. Mashor, ``Ecg signals classification based on discrete
  wavelet transform, time domain and frequency domain features,'' in \emph{2015
  2nd International Conference on Biomedical Engineering (ICoBE)}.\hskip 1em
  plus 0.5em minus 0.4em\relax IEEE, 2015, pp. 1--6.

\bibitem{tereshchenko2015frequency}
L.~G. Tereshchenko and M.~E. Josephson, ``Frequency content and characteristics
  of ventricular conduction,'' \emph{Journal of Electrocardiology}, vol.~48,
  no.~6, pp. 933--937, 2015.

\bibitem{zhao2019classification}
L.-M. Zhao, R.~Li, W.-L. Zheng, and B.-L. Lu, ``Classification of five emotions
  from eeg and eye movement signals: Complementary representation properties,''
  in \emph{9th International IEEE/EMBS Conference on Neural Engineering
  (NER)}.\hskip 1em plus 0.5em minus 0.4em\relax IEEE, 2019, pp. 611--614.

\bibitem{song2018eeg}
T.~Song, W.~Zheng, P.~Song, and Z.~Cui, ``{EEG} emotion recognition using
  dynamical graph convolutional neural networks,'' \emph{IEEE Transactions on
  Affective Computing}, 2018.

\bibitem{Chen2017A}
J.~Chen, B.~Hu, Y.~Wang, Y.~Dai, Y.~Yao, and S.~Zhao, ``A three-stage decision
  framework for multi-subject emotion recognition using physiological
  signals,'' in \emph{IEEE International Conference on Bioinformatics \&
  Biomedicine}, 2017.

\bibitem{kim2012robust}
J.~Kim and C.~D. Scott, ``Robust kernel density estimation,'' \emph{Journal of
  Machine Learning Research}, vol.~13, no. Sep, pp. 2529--2565, 2012.

\end{thebibliography}
%
%
%

%

\begin{IEEEbiography}[{\includegraphics[width=1in,height=1.25in,clip,keepaspectratio]{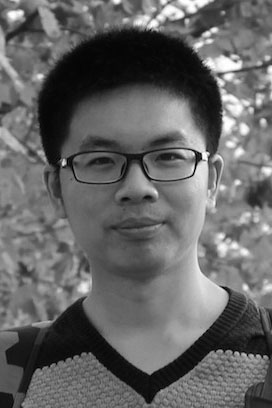}}]{Wei Liu}
received his bachelor's degree
in Automation Science from the School of Advanced Engineering, Beihang University, Beijing, China, in 2014. He is currently pursuing his Ph.D degree in
Computer Science from the Department of Computer
Science and Engineering, Shanghai Jiao Tong University,
Shanghai, China.\par 
His research focuses on affective computing,
brain-computer interface, and machine learning.
\end{IEEEbiography}

\begin{IEEEbiography}[{\includegraphics[width=1in,height=1.25in,clip,keepaspectratio]{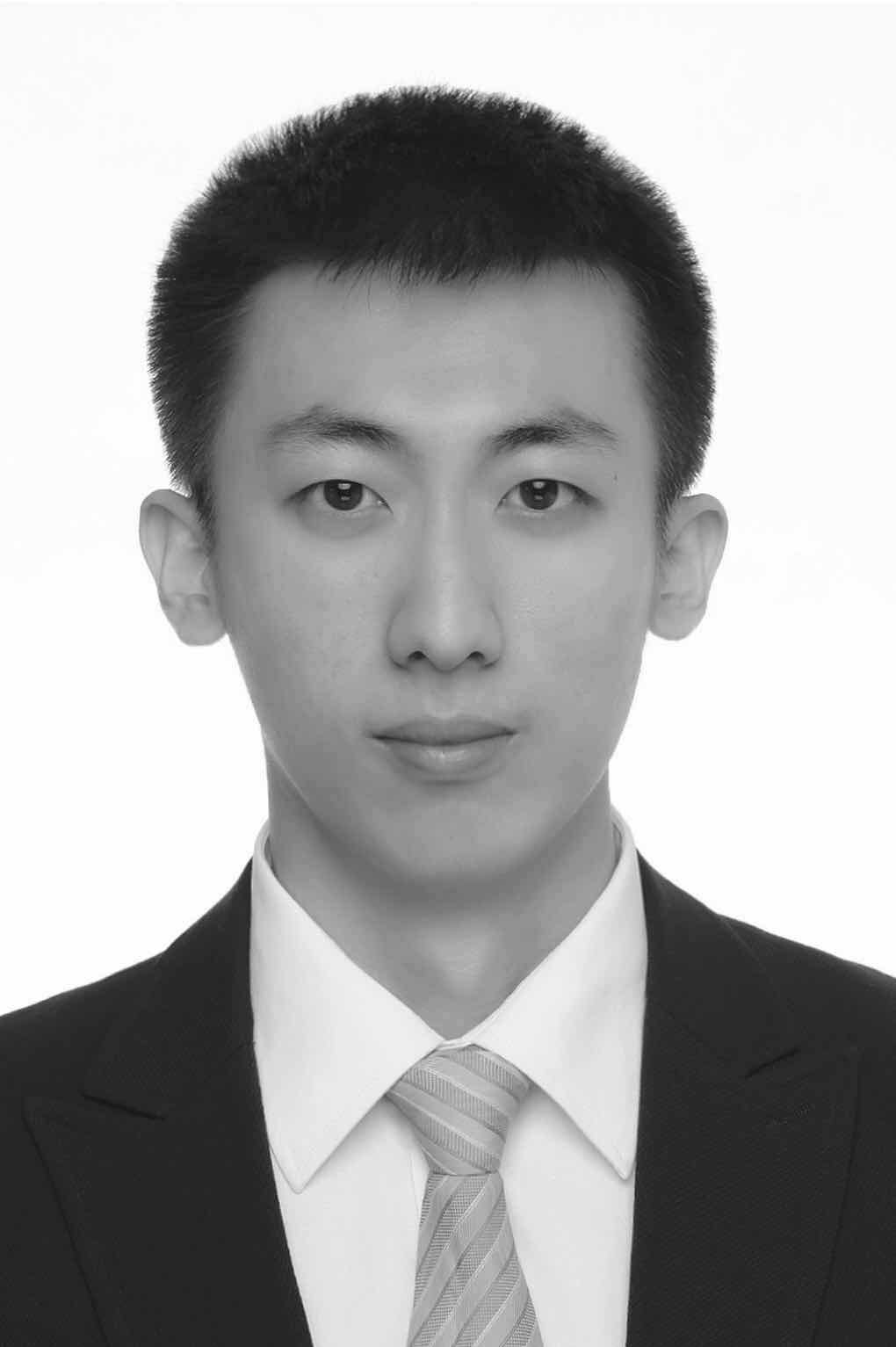}}]{Jie-Lin Qiu}
is an undergraduate student at the Department of Electronic Engineering, Shanghai Jiao Tong University, Shanghai, China. \par 
His research interests lie in the general area of machine learning, particularly in deep learning and reinforcement learning, as well as their applications in affective computing, brain-machine interfaces, computer vision, and robotics.
\end{IEEEbiography}

\begin{IEEEbiography}[{\includegraphics[width=1in,height=1.25in,clip,keepaspectratio]{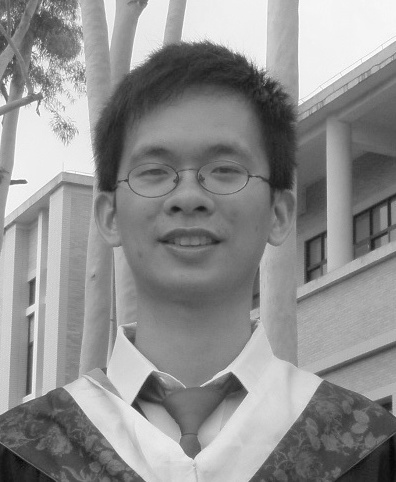}}]{Wei-Long Zheng}
(S'14--M'19) received his bachelor's degree in Information Engineering with the Department of Electronic and Information Engineering, South China University of Technology, Guangzhou, China, in 2012. He received his Ph.D. degree in Computer Science from the Department of Computer Science and Engineering, Shanghai Jiao Tong University, Shanghai, China, in 2018. Since 2018, he has been a research fellow with the Department of Neurology, Massachusetts General Hospital, Harvard Medical School, Boston, MA, USA. He received the IEEE Transactions on Autonomous Mental Development Outstanding Paper Award from the IEEE Computational Intelligence Society in 2018. His research focuses on affective computing, brain-computer interaction, machine learning and clinical healthcare.
\end{IEEEbiography}


\begin{IEEEbiography}[{\includegraphics[width=1in,height=1.25in,clip,keepaspectratio]{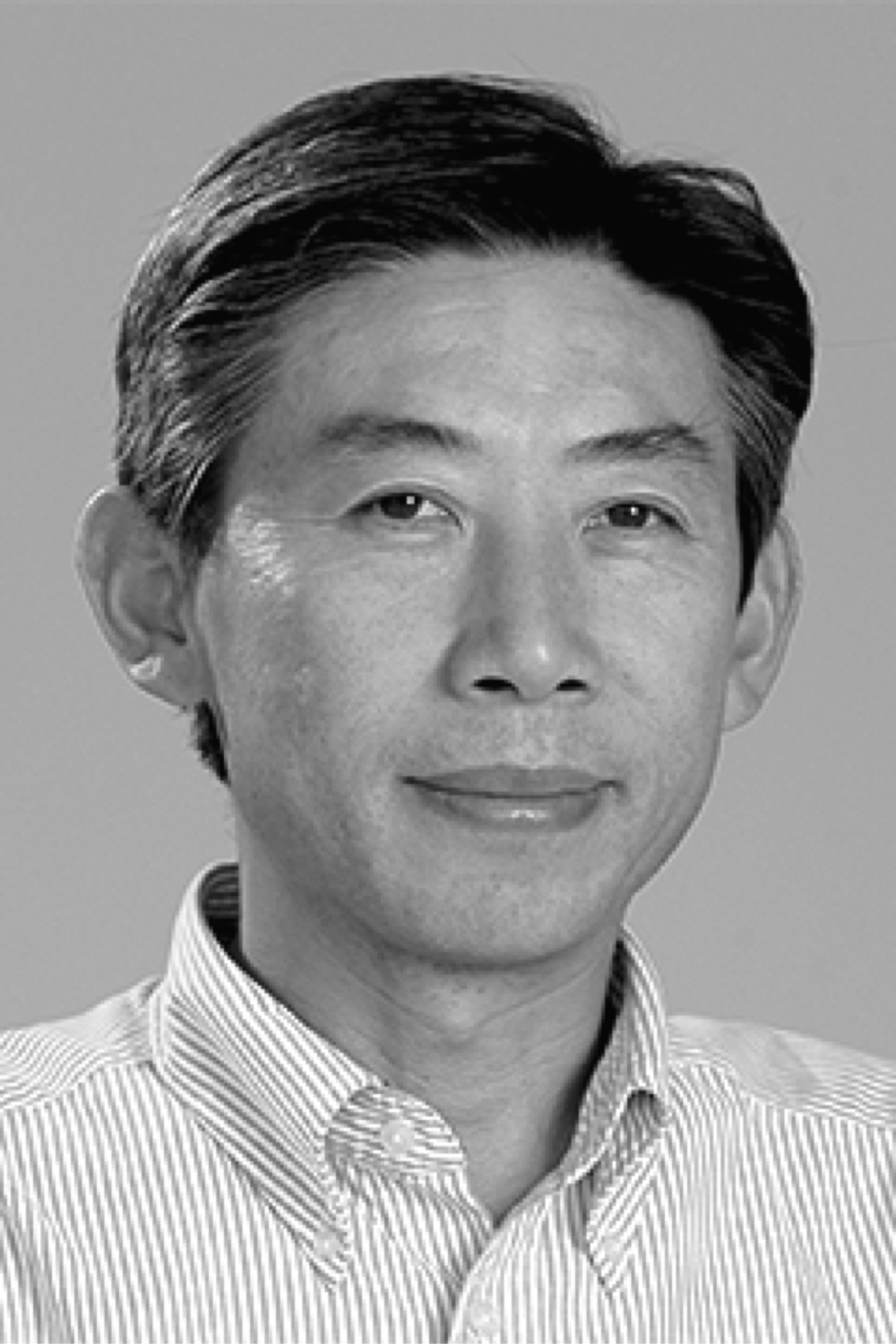}}]{Bao-Liang Lu}
(M'94--SM'10)
received his B.S. degree in Instrument and Control Engineering from the Qingdao University of Science and Technology, Qingdao, China, in 1982; his M.S. degree in Computer Science and Technology from Northwestern Polytechnical University, Xi’an, China, in 1989; and his Dr.Eng. degree in Electrical Engineering from Kyoto University, Kyoto, Japan, in 1994.\par 
He was with the Qingdao University of Science and Technology from 1982 to 1986. From 1994 to 1999, he was a Frontier Researcher with the
Bio-Mimetic Control Research Center, Institute of Physical and Chemical Research (RIKEN), Nagoya, Japan, and a Research Scientist with the RIKEN Brain Science Institute, Wako, Japan, from 1999 to 2002. Since 2002, he has been a Full Professor with the Department of Computer Science and Engineering, Shanghai Jiao Tong University, Shanghai, China. He received the IEEE Transactions on Autonomous Mental Development Outstanding Paper Award from the IEEE Computational Intelligence Society in 2018. 
His current research interests include brain-like computing, neural networks, machine learning, brain-computer interaction, and affective computing.\par 
Prof. Lu is currently a Board Member of the Asia Pacific Neural Network Society (APNNS, previously APNNA) and a Steering Committee Member of the IEEE Transactions on Affective Computing. He was the President of the Asia Pacific Neural Network Assembly (APNNA) and the General Chair of the 18th International Conference on Neural Information Processing in 2011. He is currently the Associate Editor of the IEEE Transactions on Cognitive and Development Systems.
\end{IEEEbiography}




\end{document}